\documentclass[11pt]{article}
\usepackage{latexsym}
\usepackage{color}
\usepackage{amssymb, graphicx, amsmath}
\usepackage{amsthm}
\usepackage{wrapfig}
\usepackage{subfigure}
\usepackage{graphicx}
\usepackage{url}
\usepackage{CJK,graphicx}
\usepackage{algorithm}
\usepackage{makecell}
\usepackage{multirow}
\usepackage{algpseudocode}
\usepackage{subfigure}
\usepackage{float}
\usepackage{appendix}

\usepackage{algorithm}

\usepackage{amssymb}% http://ctan.org/pkg/amssymb
\usepackage{pifont}% http://ctan.org/pkg/pifont

\usepackage{amsfonts,amsmath,amsthm,amsfonts,amssymb}
\usepackage{graphics,epsfig,graphicx,subfigure}
\usepackage[table]{xcolor}
\usepackage{helvet,epsfig,graphicx,subfigure}
\usepackage[table]{xcolor}

\usepackage{textcomp}

\usepackage{tcolorbox}
\usepackage[normalem]{ulem}

\theoremstyle{definition}

\theoremstyle{remark}

\numberwithin{equation}{section}

\usepackage{algorithm}
\usepackage{tikz}
\usepackage{algpseudocode}
\usepackage{amsmath}  
\usepackage{booktabs,array,xcolor,colortbl,multirow,epsfig,graphicx,subfigure,amsmath,amsthm,amsfonts,amssymb,bm}

\begin{document}
%\begin{CJK*}{GBK}{song}

\title{Learning Structured Communication for Multi-agent \\ Reinforcement Learning}

\author{Junjie~Sheng\thanks{School of Computer Science and Technology, East China Normal University, Shanghai 200092, China. (E-mail: 52194501003@stu.ecnu.edu.cn)}
\and Xiangfeng~Wang\thanks{School of Computer Science and Technology, East China Normal University, Shanghai 200062, China. (E-mail: xfwang@sei.ecnu.edu.cn)}
\and  Bo~Jin\thanks{School of Computer Science and Technology, East China Normal University, Shanghai 200062, China. (E-mail: bjin@cs.ecnu.edu.cn)}
\and Junchi~Yan\thanks{Department of Computer Science and Engineering, Artificial Intelligence Institute, Shanghai Jiao Tong University, Shanghai 200240, China. (E-mail: yanjunchi@sjtu.edu.cn)}
\and Wenhao~Li\thanks{School of Computer Science and Technology, East China Normal University, Shanghai 200092, China. (E-mail: 52194501026@stu.ecnu.edu.cn)}
\and Tsung-Hui~Chang\thanks{School of Science and Engineering, The Chinese University of Hong Kong (Shenzhen), Shenzhen, China. (Email: tsunghui.chang@ieee.org)}
\and Jun~Wang\thanks{School of Computer Science and Technology, East China Normal University, Shanghai 200062, China. (E-mail: jwang@sei.ecnu.edu.cn)}
\and Hongyuan~Zha\thanks{School of Computational Science and Engineering, College of Computing, Georgia Institute of Technology, USA. (E-mail: zha@cc.gatech.edu)}
%School of Computer Science and Technology, East China Normal University, Shanghai 200062, China. (E-mail: zha@cs.ecnu.edu.cn)}
%~\IEEEmembership{Member,~IEEE,}
% <-this % stops a space
%\thanks{J. Wang, X. Wang, B. Jin, and H. Zha are with the Department of Computer Science and Technology, East China Normal University, Shanghai 200092, China. J. Yan is with the Department of Computer Science and Engineering, Shanghai Jiaotong University, Shanghai 200240, China. W. Zhang is with Tecent, Shenzhen 518000, China.}
}

\date{\today}

\maketitle

\begin{abstract}
{
This work explores the large-scale multi-agent communication mechanism under a multi-agent reinforcement learning (MARL) setting. We summarize the general categories of topology for communication structures in MARL literature, which are often manually specified. Then we propose a novel framework termed as Learning Structured Communication ({\bf{LSC}}) by using a more flexible and efficient communication topology. Our framework allows for adaptive agent grouping to form different hierarchical formations over episodes, which is generated by an auxiliary task combined with a hierarchical routing protocol. Given each formed topology, a hierarchical graph neural network is learned to enable effective message information generation and propagation among inter- and intra-group communications. In contrast to existing communication mechanisms, our method has an explicit while learnable design for hierarchical communication. Experiments on challenging tasks show the proposed LSC enjoys high communication efficiency, scalability, and global cooperation capability.
}
%(MAgent and StarCraft2) 
\end{abstract}

%\mhcomment{learning to communicate, not learning to communication.}

%\vspace{-0.8cm}
\section{Introduction}%\vspace{-0.1cm}
Reinforcement learning (RL) has achieved remarkable success in solving single-agent sequential decision problems under interactive and complicated environments, such as games~\cite{Mnih2015HumanlevelCT,Silver2016} and robotics~\cite{Lillicrap2016ContinuousCW}.
In many real-world applications such as intelligent transportation systems~\cite{adler2002cooperative} and unmanned systems \cite{semsar2009multi}, not only one, but usually a large number of agents are involved in the learning tasks.
Such a setting naturally leads to the popular multi-agent reinforcement learning (MARL) problems, where the key research challenges include how to design scalable and efficient learning schemes under a non-stationary environment (caused by partial observation and/or the dynamics of other agents' policies) with large and/or dynamic problem dimension, and complicated uncertain relationship between agents.

%\mhcomment{there is a gap between these two paragraph; the previous paragraph talks about issues in MARL, while the next paragraph talks about communication-based MARL.}

%\mhdelete{To address these general issues, communication has been a popular technique in a way of strengthening collaboration strategy.}
%Various learning to communicate MARL methods have been devised e.g., DIAL~\cite{foerster2016learning}, CommNet~\cite{Sukhbaatar2016LearningMC}, ATOC~\cite{jiang2018learning}, {IC3Net}~\cite{singh2018learning} and TarMAC~\cite{das2019tarmac}. 

\begin{figure*}[htb!]
	\centering
	\subfigure[Fully-connected]{
	\includegraphics[width=0.16\textwidth]{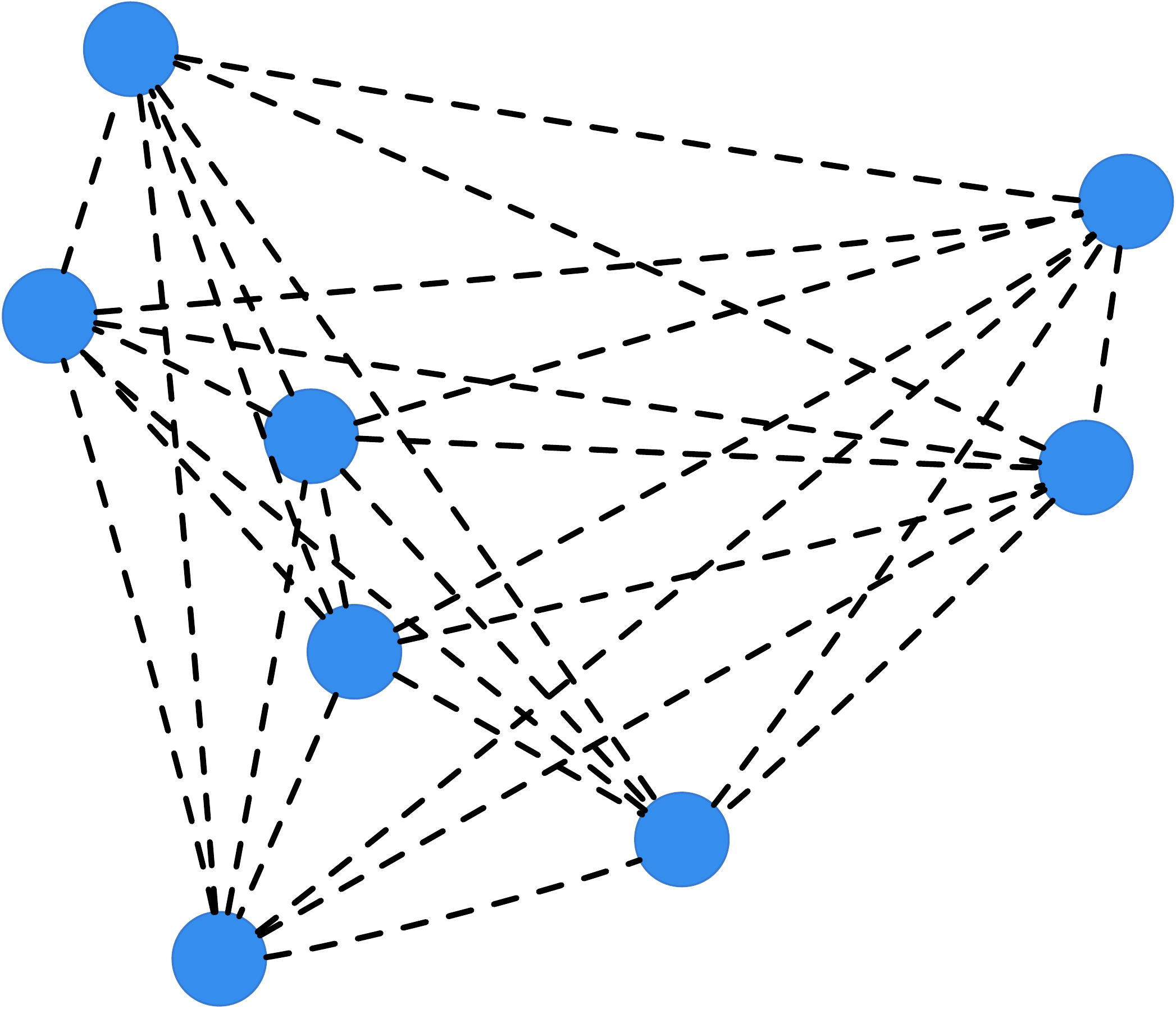}
	\label{topology-fc}
	}
	\subfigure[Star]{
	\includegraphics[width=0.16\textwidth]{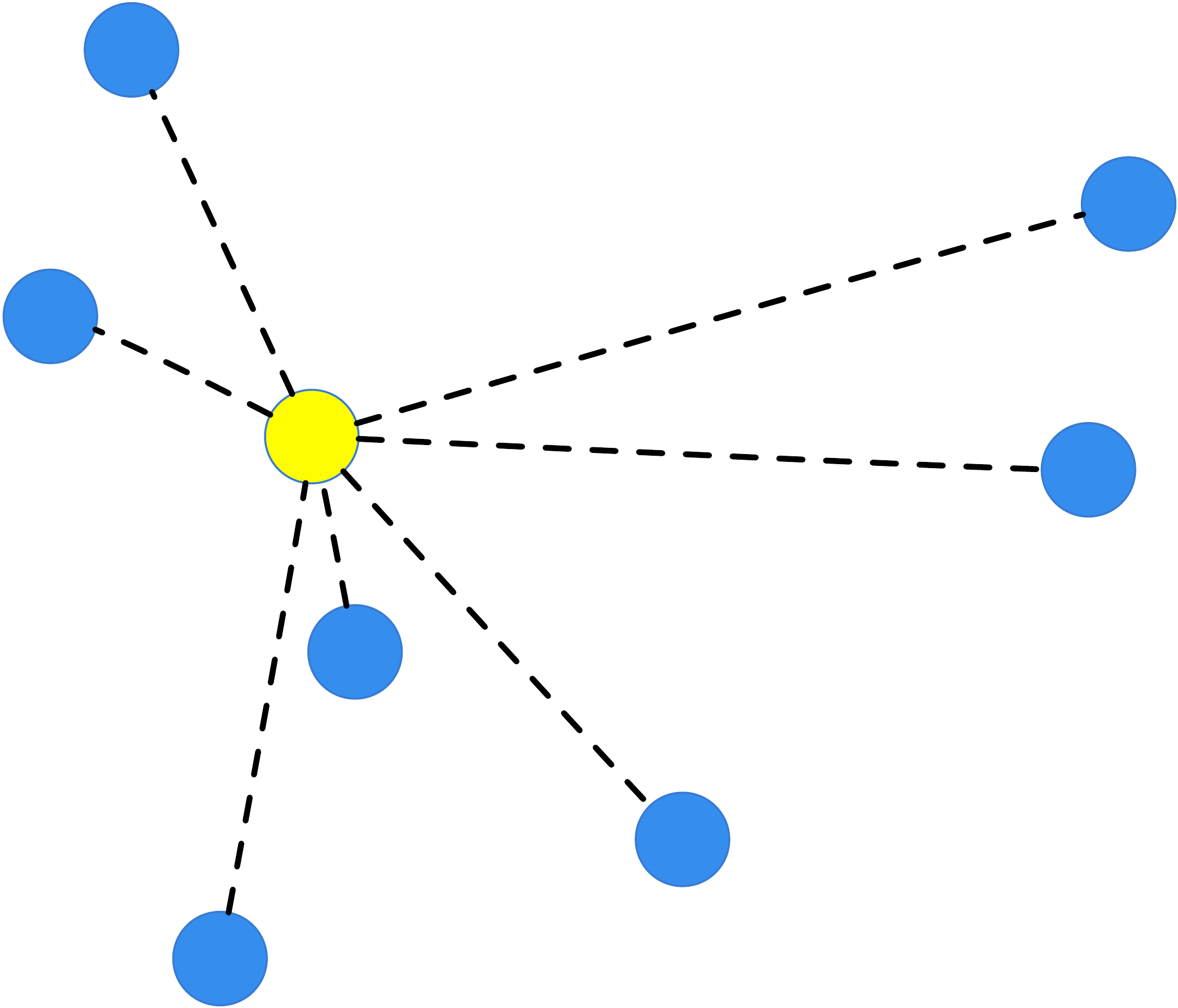}
	\label{topology-s}
	}
	\subfigure[Tree]{
	\includegraphics[width=0.16\textwidth]{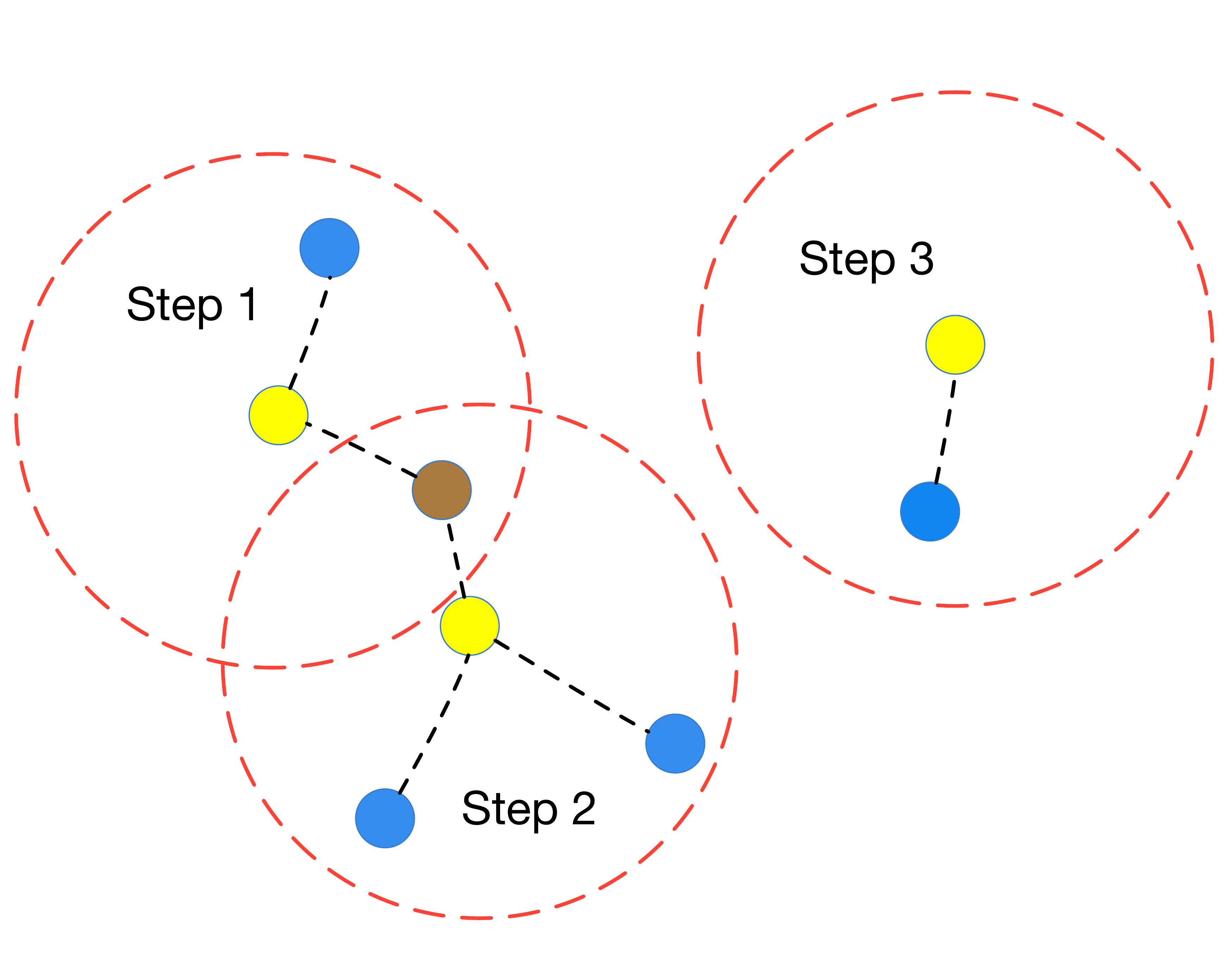}
	\label{topology-t}
	}
	\subfigure[Neighboring]{
	\includegraphics[width=0.16\textwidth]{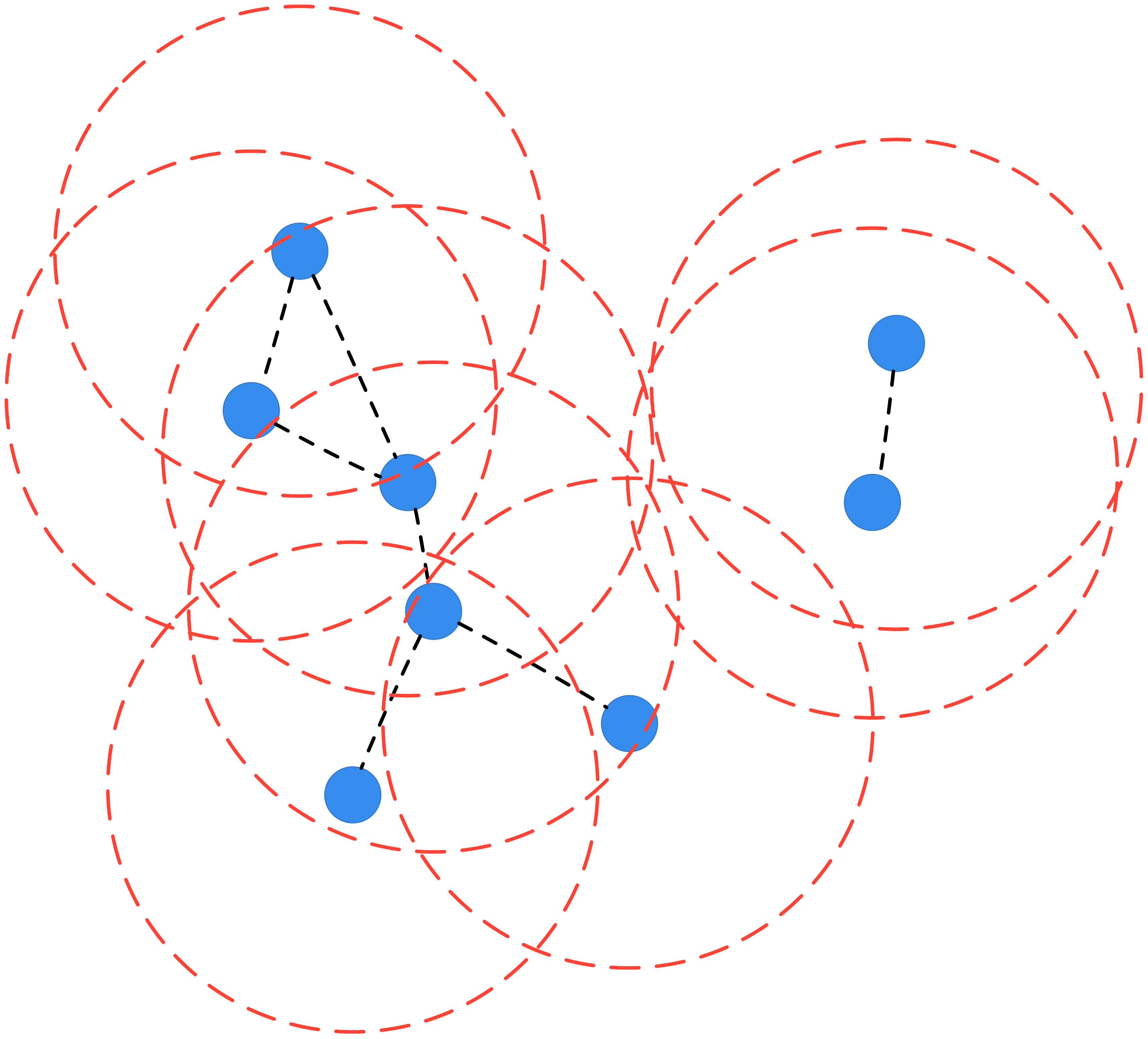}
	\label{topology-n}
	}
	\subfigure[Hierarchical]{
	\includegraphics[width=0.16\textwidth]{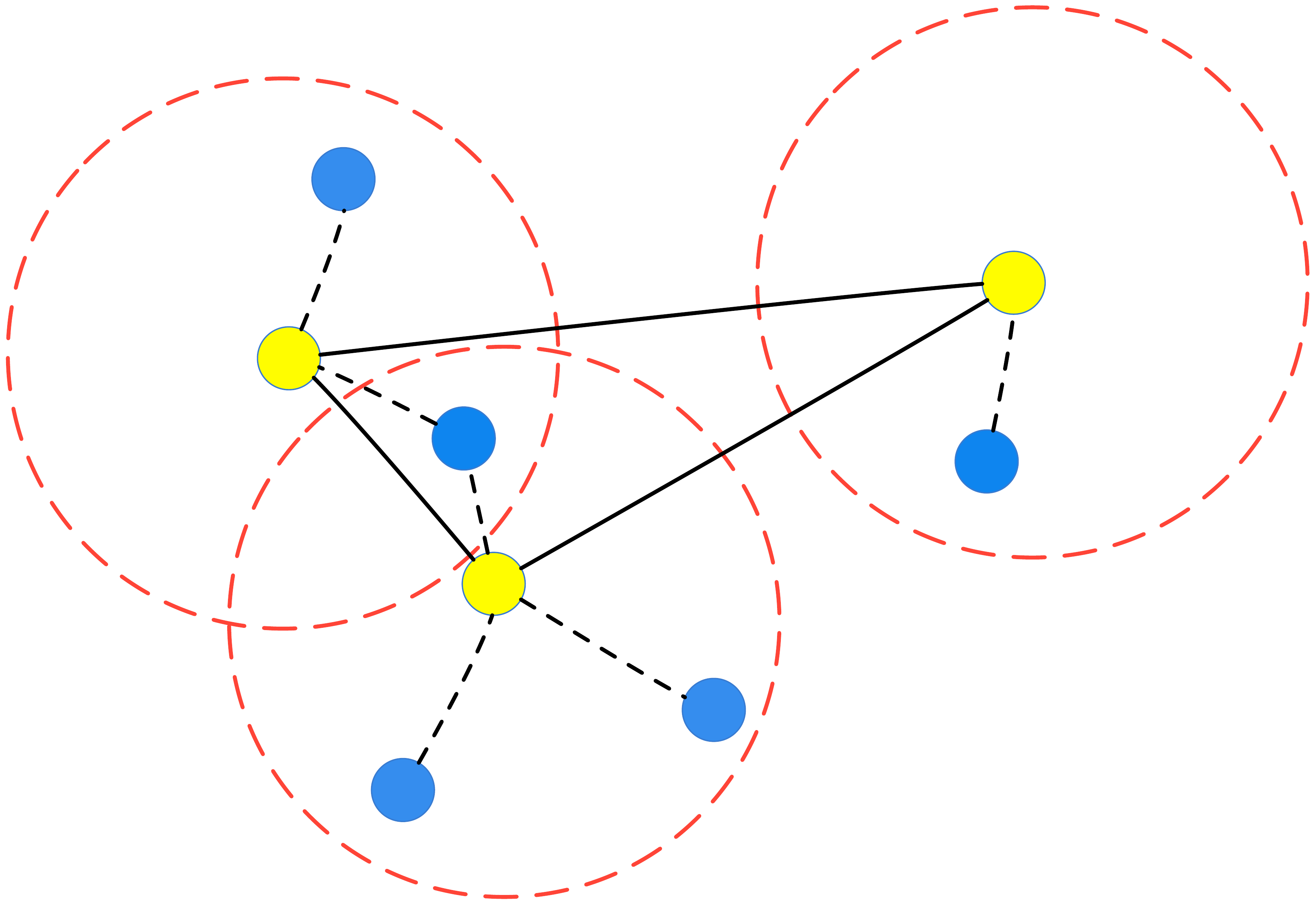}
	\label{topology-h}
	}
% 	\vspace{-6pt}
    \caption{Topology of different communication structures, and LSC falls into the hierarchical one.}
    \label{topology}
    % \vspace{-6pt}
\end{figure*}

%{\color{red}{why here??}}{\color{blue}{Last but not least, the problem of global cooperation for large-scale MARL problem remains open.}}

%From the perspective of communication efficiency, there are two important factors: communication bandwidth and communication time complexity. 
Learning to communicate effectively among agents has shown crucial to strengthen the inter-agent collaboration and ultimately improve the quality of policies learned by MARL. In this paper, we categorize the existing designs for communication topology\footnote{We interchangeably abuse the term topology and structure.} into four patterns: 
\textbf{i) Fully-connected:} DIAL~\cite{foerster2016learning}, TarMAC~\cite{das2019tarmac} and SchedNet~\cite{kim2018learning} use fully-connected communication structure (see Fig.~\ref{topology-fc}). Agents need to communicate with all the others, thus requiring high bandwidth when the number of agents is large. 
\textbf{ii) Star:} CommNet~\cite{foerster2016learning} and IC3~\cite{singh2018learning} assume star communication structure (see Fig.~\ref{topology-s}). 
All agents need to transmit messages to the virtual central agent, incurring a major communication bottleneck.
\textbf{iii) Tree:} ATOC~\cite{jiang2018learning} uses a tree communication structure. Agents only communicate with neighbors. However, communication must be allowed sequentially among groups, leading to high time complexity. 
\textbf{iv) neighboring:} DGN~\cite{Jiang2020Graph} uses neighboring communication structure. Agents communicate with neighbors concurrently to reduce communication costs.

We further analyze the above communication topology patterns by considering the accessibility and comprehension of messages for effective communication. Both the fully-connected structure and star structure ensure messages are accessible for all agents. While as discussed in ATOC~\cite{jiang2018learning}, once a large number of messages emerge concurrently, extracting valuable information would become difficult. Tree structure and neighboring structure constrain the communication to neighbors and hence is able to improve message comprehension. To achieve global accessibility, they define neighbors as $K$-nearest agents and utilize multi-round communications. However, due to the lack of pooling mechanism, DGN ~\cite{Jiang2020Graph} incurs two rounds of convolutions in communication; thus, only the information from two-hop distant agents is aggregated. %Agents can only aggregate , thus the global communication can not be adequately achieved for every agent.
In this paper, we are aimed to improve both efficient message accessibility and effective message comprehension for large-scale MARL by proposing the so-called {\bf{L}}earning {\bf{S}}tructured {\bf{C}}ommunication ({\bf{LSC}}) approach. Specifically, our LSC contains a structured communication module and a communication-based policy module. It aims to establish a hierarchical communication structure for learning the communication pattern as well as policy. 
In particular, a hierarchical communication structure (Fig.~\ref{topology-h}) is established in a distributed fashion by a cluster-based routing protocol. 
To make the structure be formed dynamically for better cooperation, an auxiliary reinforcement task is designed to learn the communication weight in the end-to-end fashion.
% We add an auxiliary task to learn the communication weight, combine with a cluster-based routing protocol to establish the hierarchical structure. 
In the hierarchical structure, agents are grouped to different groups, and every group is assigned with a high-level agent. We design an intra-inter group communication mechanism to achieve global communication efficiently. Inter-group communication can help agents to capture global information better while intra-group communication helps fine-grained message exchanges. With these two modules, our experiments show that LSC can efficiently achieve global communication efficiency. The main highlights of this paper are summarized below.

1) We summarize the four existing categories of communication topology in the MARL literature, namely i) fully-connected, ii) star, iii) tree, and iv) neighboring, which are in general manually specified and fixed. We believe this perspective is enlightening for the design of new communication topology, given the fact that it has not been well organized in the existing literature.

2) We develop a new hierarchical communication topology LSC, which differs from the existing four patterns. Our approach allows for an adaptive formation of agents by dynamically grouping agents via a reinforcement learning procedure combined with a routing protocol. The messages can be jointly extracted and propagated through both intra- and inter-group communications via a hierarchical graph neural network.  

3) Experimental results show that the proposed LSC yields promising results on public benchmarks in terms of communication efficiency, scalability, and global cooperation capability. 

To our best knowledge, the current paper is the first work about hierarchical communication learning in MARL. We note that the idea of adopting hierarchical structure learning on MARL recently appears in HAMA~\cite{hama}. 
The differences are obvious and fundamental: first, their hierarchy structure is used for learning agents' relation, but not for communication; second, their hierarchy design is fixed, other than adaptive and dynamically learned as done in this paper.

%The source code and models will be made publicly available with the final version of the released paper.

% This paper is devoted to the learning of communication structure among agents. 
% To our knowledge, this is the first work of hierarchical structured learning to communication for MARL. 
% It allows to learn communication structure adaptively instead of using predefined forms. Specifically:

% \vspace{-0.1cm}
% i) To improve scalability for a large number of agents, a hierarchical structure is devised that divides the agents into higher-level central agents and sub-level normal ones. 
% \jar{The hierarchical structure allows global communication with time }
% As such, the communication network is sparsified. 
% While it still allows for more effective global cooperation via message passing among the central agents, compared with the star/tree structures.

% \vspace{-0.1cm}
% ii) For effective communication and global cooperation, the message representation learning is deeply integrated into the information aggregating and permeating through the network, via graph neural network (GNN), which is a natural combination with the hierarchical communication structure.
% \jar{ii) We design an auxiliary task to learn the communication structure. This cn}
% \vspace{-0.1cm} 

% iii) Extensive experiments on both MAgent and MPE show our approach achieves state-of-the-art scalability and effectiveness on large-scale MARL problems.

%%%%%%%%%%%%%%%%%%%%%%%%%%%%%%%%%%%%%%%%%%%%%%%%%%%%%%%%%%%%%%%%%%%%%%%%%%%%%%%%%%%%%%%%
% \vspace{-0.3cm}
\section{Related Work}
% \vspace{-0.1cm}
% Many multi-agent reinforcement learning algorithms without communication in the inference procedure have experienced fast development. 
%We discuss related work based on how the consensus is achieved in agents, and graph neural networks for MARL.

\paragraph{Learning-for-consensus.} 
These approaches try to let agents achieve consensus and cooperation directly from local observations, whereby a centralized training and decentralized execution framework (CTDE) is often used. Methods like MADDPG~\cite{lowe2017multi}, QMIX~\cite{rashid2018qmix}, COMA~\cite{foerster2018counterfactual} and MAAC~\cite{iqbal2019actor} concatenate all the agents' observations and/or polices to obtain the state representation. 
This helps achieve better cooperation. However, the curse of dimensionality occurs when a large number of agents are present. 
HAMA~\cite{hama} adopts a hierarchical graph attention network to leverage the group relationships.
However, the groups are clustered by predefined rules, which is not feasible for complex scenarios.

% Recent works like adopt a centralized training and decentralized execution framework. 
% All agents' local observations and actions are considered to improve the learning stability. 
% These algorithms are generally not suitable for large-scale case due to explosive growing number of agents.

% Communication-based MARL algorithms have been showed effective for large-scale agent cooperation. 
% Earlier works assume that all agents need to communicate with each other. 
% \paragraph{Communication Topology in MARL.}
{\bf Learning-for-communication.}
In these approaches, the agents aim to achieve consensus and cooperation through communications. Agents need to learn to communicate with others and process the received messages to enhance collaboration.
As mentioned, the communication topology of the existing methods can be categorized as i) fully-connected (FC); ii) star; iii) tree; and iv) neighboring.

Fully-connected structures assume that each agent communicates with all the other agents. DIAL~\cite{foerster2016learning} learns what to communicate by back-propagating all the other agents' gradients to the message generation network. SchedNet~\cite{kim2018learning} learns a weight-based scheduler to determine the communication priority based on DIAL, but the way of using the communication bandwidth is not scalable. Star structures assume agents only communicate with the single central agent as the hub. CommNet~\cite{foerster2016learning} aggregates all the agents' hidden states as the global message, thus can only be applied to cooperative scenarios. Extended from CommNet, IC3~\cite{singh2018learning} adds a communication gate to decide whether the agents to communicate. However, letting one agent handle all the messages in the star network cause a bottleneck at the central agent, both in communication bandwidth and information extraction. Tree and neighboring structures constrain communication to neighbors, thus avoids the single-point bottleneck issue. The $K$-nearest neighbor mechanism is often used to define neighbors. However, agents can distribute unevenly, and thereby choosing a good $K$ is sometimes not easy in practical scenarios. ATOC~\cite{jiang2018learning} adopts tree structures, whereby each group in a chain (thus not hierarchical) performs communication sequentially. Although the inter-group communication can be achieved by the intersection of two groups, the large time complexity would be unbearable for real-time systems. To address the aforementioned difficulties, DGN~\cite{Jiang2020Graph} uses neighboring structured communication together with the graph convolution network (GCN). Multiple rounds of communications are adopted to enlarge the receptive field. As a common issue in GCN, shallow GCN without pooling layers can hardly explore rich global information as discussed in H-GCN~\cite{ying2018hierarchical}

% \jar{DIAL~\cite{foerster2016learning} assumes a fully connected communication and learn what to communicate by backpropagating all other agents' gradients to message generation network. However the fully connected communication structure is inefficient for large scale environments. CommNet~\cite{Sukhbaatar2016LearningMC} and IC3~\cite{singh2018learning} both assume a star communication and take the mean of all the hidden state as message, while IC3 uses a communication gate to decide whether to send the message or not. The star communication structure require large communication range and can hard to extract fine-grained information when the number of agent grows large. ATOC~\cite{jiang2018learning} and DGN~\cite{Jiang2020Graph} constrain communication only happen on K-Nearest neighbour to ease the information extraction. Although the K-nearest communication ensure no agent are disconnected, agents distribute unbalanced in practical scenarios. Thus k can hard to be chosen especially for large and complex scenarios. Moreover, ATOC assumes a tree communication(every group should communicate sequentially) which lead to time inefficient. DGN use neighboring communication to gain better communication efficiency, however adequate global information can hard to achieved by shallow GCN. }

%The spectral approaches like DGCN~\cite{zhuang2018dual} use spectral graph theory to define filters. The spatial approaches aggregate neighbors embeddings directly. 
{\bf MARL with Graph Neural Network (GNN).} GNN is powerful in extracting relations among entities, with emerging applications in MARL. RFM~\cite{tacchetti2018relational} designs an auxiliary action prediction task (predict other agents' actions) with graph networks~\cite{battaglia2018relational}, which can help agents learn interpretable intermediate representations.
%However, for partial observable scenarios, using local observation to predict actions can be challenging.
MAGNet~\cite{aleks2018deep} uses heuristic rules to learn the relevant graph to help actor and critic learning.
%uses to extract relations among entities and further uses it to predict other agents' actions. However, for complex systems, appropriate design rules can be difficult.
DGN~\cite{Jiang2020Graph} learns the GCN together with the relation kernel by minimizing the TD error, which can be applied to dynamic multi-agent RL problems.
%However, lacking the pooling layer and shallow GCN can be hard to capture adequate global information.
HAMA~\cite{hama} adopts a hierarchical graph attention network based on a pre-defined hierarchical graph to help agents capture interrelations.
%, it can be helpful to capture global information. 
The pre-defined and fixed group scheme used in HAMA limits its adaptability in dynamic scenarios.

In this paper, we target at learning both the underlying topology (with a hierarchy prior design) and the top-layered message extraction as well as the propagation process via GNNs. The message communication mechanism, the underlying topology as well as the way of using GNN in this paper all are novel and different from the existing works.
\section{LSC: Learning Structured Communication}
% \vspace{-0.1cm}
%This section presents our learning structured communication framework, which can support dynamic organization of communication network and message representation.
\subsection{Preliminaries and Overview}
{\bf Partial Observable Stochastic Games (POSG).}
Agents learn policies by maximizing cumulative rewards via interacting with environment and other agents. POSG can be characterized as a tuple {$\left\langle {\cal{I}}, {\cal{S}}, b^{0}, \mathcal{A}, \mathcal{O}, {\cal{P}}, \mathcal{P}_{e}, \mathcal{R} \right\rangle$} where $\mathcal{I}$ denotes the set of agents indexed from $1$ to $n$; ${\cal{S}}$ is the finite set of states; $b^{0}$ represents the initial state distribution and $\mathcal{A}$ denotes the set of joint actions. $A_i$ is the action space of agent $i$, $\mathbf{a}=\langle a_1,\cdots,a_n\rangle$ denotes a joint action; $\mathcal{O}$ denotes the joint observations and $O_i$ is the observation space for agent $i$, $\mathbf{o}=\langle o_1,\cdots,o_n\rangle$ denotes a joint observation; ${\cal{P}}$ denotes the Markovian transition distribution with $P\left(\tilde{s}, \mathbf{o} \big| s, \mathbf{ a } \right)$ being the probability of state $s$ transiting to $\tilde{s}$ with result $\mathbf{o}$ after taking action $\mathbf{a}$. $\mathcal{P}_{e}(\boldsymbol{o}|s)$ is the Markovian observation emission probability. $\mathcal{R}:\cal{S} \times \mathcal{A} \rightarrow$ ${\mathbb{R}}^{n} $ means the reward function.
$\mathbf{r} = \langle r_1,\cdots,r_n\rangle$ denotes the joint reward each agent. 
%with ${\matohbb{R}}^n$ denote a real coordinate space of $n$ dimensions.
The overall task of MARL can be solved by proper objective modeling, which may indicate, e.g., cooperative, competitive, or mixed relationship among agents. 

{\bf Deep Q-Learning.}
Deep $Q$-Network (DQN)~\cite{Mnih2015HumanlevelCT} is popular in deep RL as it is one of the few RL methods applicable to large-scale MARL. In each step, each agent observes state $s$ and takes an action $a$ based on policy $\pi$. It receives reward $r$ and next state $\tilde{s}$ from the environment. To maximize the cumulative reward after step $t$, $R_t = \sum_{k=t}{r_k}$, DQN learns the action-value function $Q^{\pi}(s, a) = \mathbb {E}_{s\sim \mathcal{P}, a\sim\pi(s)} \left[ R _ { t } | s _ { t } = s , a _ { t } = a \right]$ by minimizing $\mathcal { L } ( \theta ) = \mathbb { E } _ { s , a , r , \tilde{s}} \left[ \tilde{y} - Q ( s , a ; \theta ) \right]$, where  $\tilde{y} = r + \gamma \max _ {\tilde{a}} Q \left( \tilde{s} , \tilde{a} ; \theta \right)$. The agent follows $\epsilon$-greedy policy: select the action that maximizes the $Q$-value with probability $1$-$\epsilon$ or randomly. Independent Deep $Q$-Learning (IDQN)~\cite{Tampuu2017} extends DQN by ignoring other agents for the POSG. Each agent learns a $Q$-function $Q^a(u^a|s;\theta^a)$ based on its own observation and reward. Our LSC extends DQN with hierarchical communication.

{\bf Proposed Approach Overview.} LSC takes the aforementioned formulation of POSG with the communication mechanism taken into account. Every agent needs to learn both action and communication policies. 
As discussed in MFRL~\cite{Yang2018MeanFM}, the Q-Learning family is more stable than on-policy methods in large-scale MARL. 
Hence, we choose DQN to learn the action policy with communication and form the global state perception.

\begin{figure}[tb!]
	\centering
	\includegraphics[width=0.8\textwidth]{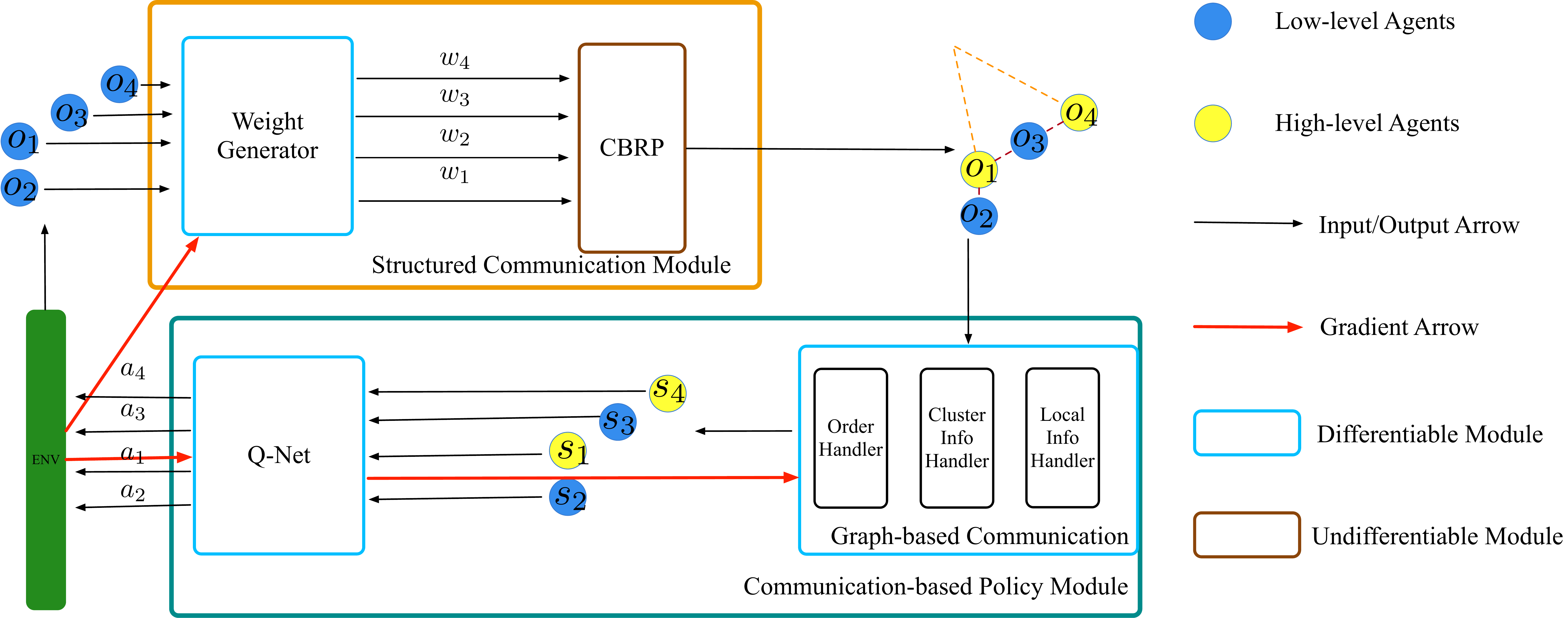}
% 	\vspace{-20pt}
    \caption{LSC with {\em{Structured Communication Module}} and {\em{Communication-based Policy Module}}, where $s_i$, $o_i$, $a_i$ and $w_i$ denote state (global perception), observation, action and importance weight of agent $i$, respectively. The former module uses partial observation to establish the communication structure. The latter employs Graph-based communication and $Q$-Network to extract communication content and produces collaboration policies respectively. based on established communication structure.} %\mhcomment{``non-differentiable"; symbols and notations in the graph not defined, e.g., CBRP, ENV, order handler, cluster info handler, etc.}}
    \label{arch}
    % \vspace{-0.5cm}
\end{figure}

To construct a hierarchical communication structure, LSC designs a flexible two-level communication topology, where agents are dynamically divided into high-level agents and low-level agents, indicated by the yellow and blue points in Fig.\ref{topology-h}. The high-level agents are in charge of forming global perception and coordinating low-level agents in their group. The low-level agents need to convey the local information to high-level agents. LSC includes two key modules: i) structured communication module and ii) {communication-based policy module}, as shown in Fig.~\ref{arch}. The first module aims to establish the dynamic hierarchical structured communication topology in a distributed fashion, while the second module contains the GNN-based communication extraction and $Q$-network components.

% \vspace{-0.4cm}
\subsection{Structured Communication Module}
% \vspace{-0.1cm}
\label{subsec:CSNM}
\begin{figure*}
	\centering
	\includegraphics[clip=true, width=\textwidth]{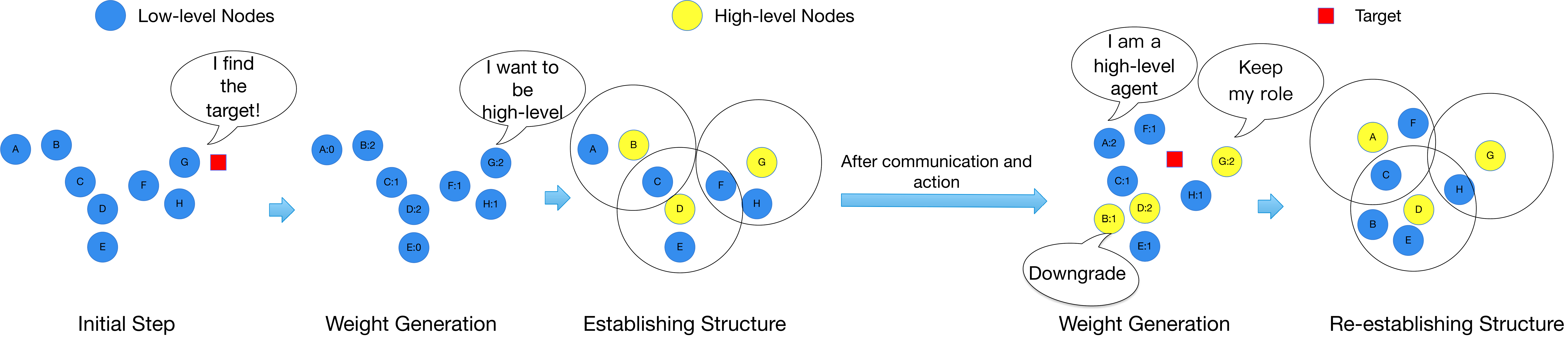}
% 	\vspace{-20pt}
    \caption{Procedure for dynamically establishing a hierarchical communication structure. Each agent determines its communication weight based on partial local observation. Agent `G' finds the target (red square), then it has a higher weight $2$ in the weight generation step. Then at the structure establishing step, it is elected as a high-level agent. After communication and actions, agents' positions may change. Each agent needs to re-generate the communication weight and decide to keep or change their communication roles. For instance, `B' gets a lower weight with a high weight agent `D' nearby. Then `D' will downgrade its role at the structure establishing step.
    %\mhcomment{not sure what the red part means}.
    %the Cluster Based Routing Protocol (CBRP) \citet{cbrp} is introduced to dynamically establish a hierarchical communication structure. 
    % {\bf Right}: The importance weight generation step and network construction step will be repeated iteratively. 
    % After communication and action procedures, agents will generate their new communication importance weights, and determine to keep or change their roles respectively. 
    % Further, the structured communication network will be re-established.
    }
    %after communication and action.}
    \label{cluster}
\end{figure*}

The structured communication module is designed by three principles:
1) agents in the same group are more likely to understand and cooperate inner group;
2) high-level agents are more likely to capture the global perception through the exchanged messages; 
3) high-level agents are distributed sparsely to lower the communication cost. 
According to ATOC~\cite{jiang2018learning} and DGN~\cite{Jiang2020Graph},
nearby agents are more likely to understand each other and form cooperation.
Thus, we use the local geometrical relationship and the policy performance as our guide to establishing the hierarchical structure, as shown in Fig. \ref{cluster}.
% (similar observation eases the difficulty to understand and the action taken by neighboring agents are more influential)
% the hierarchical structure is naturally formed by first finding some high-level agents and then assigning them to a few groups by their neighborhood.
% Moreover, agents can achieve higher performance with better global perception. 
% Thus, we can use some reward to guide the selection of high-level agents. 
% For sparsity and simplicity, the high-level agent in each group should satisfy two conditions: 
% i) higher weight (by certain means) than the surrounding agents; 
% ii) no other high-level agent in its reception field.

%take the agents neighbouring high-level agents as a group. 
%neighboring no high-level agents can be high-level agents.}

% Specifically, we add an auxiliary task with a communication protocol. The state and reward design are the same as the original task, while the action space is the weight of communication.
% The structured communication network module takes the role of establishing a hierarchical communication network that will be employed in the communication-based policy module.
Specifically, two sub-modules are included: the weight generator and the Cluster-Based Routing Protocol (CBRP). 
The weight generator sub-module aims to determine the importance of communication for each agent automatically. 
It is modeled by a neural network $f_{wg}:o_i \rightarrow w_i$, where the weight $w_i$ can measure the confidence of an agent to become high-level. 
Further, the CBRP sub-module employs the weights of all agents $\mathbf{w}$ and considers the local geometry to construct the hierarchical communication network. 
The CBRP sub-module can be implemented in a distributed fashion, leading to a distributed election of high-level agents. 
This advantage ensures the applicability of LSC to large-scale scenarios, which is demonstrated in the experiments.

The CBRP method~\cite{cbrp} is a typical method for establishing a hierarchical routing structure.
It takes a hyper-parameter cluster radius $d$ as the basis to establish structure, and we denote the agent's perceptive field as the area within cluster radius.
%The framework of the CBRP sub-module can be found in Figure xx.
%The key idea for CBRP is that each agent will check whether central agents exist or agents larger weight $w$ in its receptive area.
Each low-level agent checks whether other agents have larger weights or contain high-level agents within its receptive area.
If no such agent is found, this agent is elected as a high-level agent; otherwise, it keeps as a low-level agent.
Meanwhile, each high-level agent checks whether other high-level agents exist in its receptive field.
If no such agent is found or the founded high-level agents' weights are smaller than its weights, it keeps as a high-level agent; otherwise, it downgrades to a low-level agent.
% If no such agent is found, the agent will keep as a high-level agent; otherwise, it will downgrade to a low-level agent.
After a sufficient number of rounds, the hierarchical structure would be established with sparsity: no high-level agent is included in other high-level agents' receptive fields, which benefits communication efficiency.
All agents are separated into groups with one high-level agent as the group leader.
The overall hierarchical communication network is thus established: connecting high-level agents across groups, and connecting each low-level agent to its high-level agent.
% The key idea for CBRP is that each agent will check whether central agent or agent that has larger weight $w$ exists in its receptive area.
% The agent will become a central agent if no above agent is found, else it will keep its own role. With enough checking steps, each agent will either be an central agent or in some central agents' receptive. All agents can be separated into several groups with each central agent as the group leader. The overall hierarchical structured communication network further will be established by fully connecting all central agents from different groups and connecting the agents in each group to their central agent.
%If the answer is no, then it will elect itself to be one of the central agents. Else it will choose to mute. After multi-step checking, every agent will either in central agents' receptive or be the central agents. Every agent here can be considered as a node in a graph. Thus we have two types nodes: a) central nodes b) ordinary nodes.  Every central node connects to all other central nodes and ordinary nodes in its receptive. By far the hierarchical structure gets established.
% It is worth noting that the CBRP sub-module is not differentiable, which means the gradients cannot be back-propagated from communication-based policy module.
% However the connection between these two modules is still strong: different weight generators will lead to different hierarchical structures, which will further cause significant diverse performance of the communication-based policy.

% These two modules are strongly connected in our LSC.
A naive way of designing a weight generator is to set a fixed weight for all the agents simply. 
However, improper weights would result in a poor hierarchical structured communication network, which further causes the diverse performance of the communication-based policy.
The experimental results also suggest that the choice of weights has a non-negligible influence on the performance, which motivates us to train these two modules end-to-end.

However, the CBRP sub-module is not differentiable, which means that the gradients cannot be back-propagated from the communication-based policy module to the weight generator sub-module.
Therefore, we introduce an auxiliary RL task for weight generating, where the action for each agent is weight choosing with the same observation and reward of the original task, as well as practical constraints. 
Hence, we can have a close-loop task-driven communication weight generating manner.
%By considering the communication-based policy as an \textit{extra} unobservable part of the environment, the action for the auxiliary task is weight choosing.
% Each agent takes its weight as an action by treating the communication-based policy module as an \textit{extra} unobservable part of the environment and receiving the same reward as the main RL task in the communication-based policy module discussed below.
Specifically the weight $w$ is defined in the discrete set $\{0,1,2\}$.
IDQN is chosen to implement the weight generator for simplicity. 
% is constrained in the integer set $\{0,1,2\}$.
% The action space becomes discrete. As a result the DQN algorithm can be used again to train the weight generator.
% At this time, the weight generator can be regarded as a $Q$-value function.
The loss $\ell ( \theta^{w} )$ for the weight generator sub-module is:% defined as follows:
%\vspace{-10pt}
\begin{equation}\label{eq:loss-structure}
\ell ( \theta^{w} ) = {\mathbb{E }}_ {\mathbf{o}, \mathbf{w}, \mathbf{r}, \tilde{\mathbf{o}}} \big[ \sum_{i=1}^n ( Q_{\theta^{w}} ( o_i, w_i) - y_i ) ^ { 2 } \big].
\end{equation}
where $y_i = r_i + \gamma \max_{\tilde{w}_i} Q_{\theta^{w}} (\tilde{o}_i, \tilde{w}_i)$, and $r_i$ denotes the reward for agent $i$.

%\jar{With the two sub-modules: weight generator and CBRP, the hierarchical communication structure would be established well.}

%\vspace{-10pt}
\subsection{Communication-based Policy Module}

\begin{figure}[tb!]
	\centering
	\includegraphics[clip=true,width=0.8\textwidth]{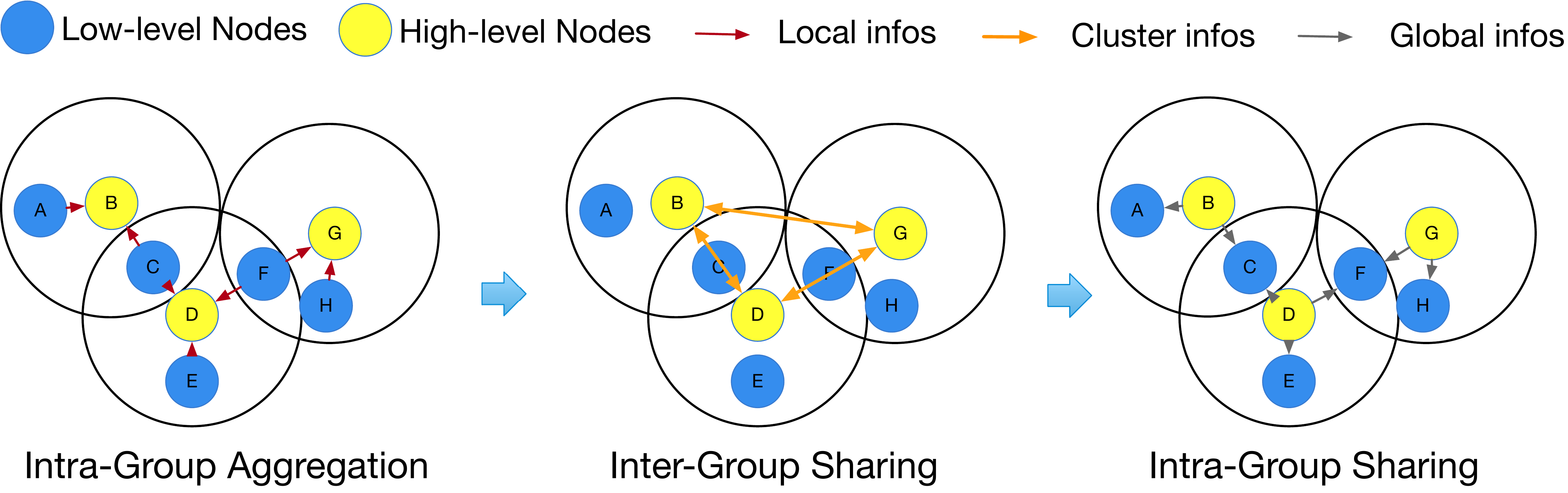}
% 	\vspace{-20pt}
    \caption{Intra-inter group communication. 
    %Each node denotes an agent.
    The edge embedding is considered as the communication message between agents.
    %The network learning procedure properly fits the communication procedure, and effectively learn valuable messages involving the global network structure and agents relationship.
    {\bf Left}: Low-level agents transfer their valuable local embeddings to the associated high-level agents. 
    {\bf Middle}: High-level agents communicate with each other to form the global perception. 
    {\bf Right}: %Central agents broadcast embedding information to their normal agents to guide cooperation within group.
    All high-level agents broadcast embedding information to their low-level agents to establish global cooperation.}
    %\mhcomment{not clear where GNN is used in the figure and in the description, i.e., how it is helping the communication?}}
    \label{comm}
    % \vspace{-0.6cm}
 \end{figure}
 
Once the communication network topology is determined, the communication-based policy module learns the communication message and generates a global collaboration policy. The communication-based policy module consists of two sub-modules: GNN-based communication sub-module and the $Q$-Net policy sub-module. The former is used to learn the communication messages and further update overall state perceptions. The latter learns the policy based on the new state perceptions after efficient communication. 
%\jar{With the established hierarchical structure, agents are grouped with each group exists a high-level agent. 
%Thus we design a new communication mechanism(inter-intra group communication) to better utilize the hierarchical structure. }
% Different from many existing works \cite{foerster2016learning,das2019tarmac,singh2018learning}, here the agents play differently in the GNN-based communication sub-module. 
% \jar{High-level agents need to obtain global perception and dominate the agents in their driven groups}
% \textcolor{red}{High-level agents should guarantee high-level information?} and dominate the agents in their driven groups, respectively. 
% The hope is that such a structure can \textcolor{red}{ensure} the effectiveness of communication and the efficiency of intra-group and inter-group collaboration.

%we assign different roles for agents when communicating. This is very common in human society. Some individuals obtain higher-level information and guide the agents they govern, thus ensuring the effectiveness of communication and the efficiency of intra-group and group-to-group collaboration.
As illustrated in Fig.~\ref{comm}, the well-established hierarchical structured communication network can be represented by a directed grpah $({\cal{V}},{\cal{E}})$.
The node set ${\cal{V}}$ contains $N_v$ nodes, which can be divided into the high-level node set ${\cal{V}}_h$ and the low-level node set ${\cal{V}}_l$. 
For $i\in {\cal{V}}_h$, the node feature vector $v_i$ includes the embedding feature $v^n_i$, the high-level node feature $v^h_i$ and the global feature $v^g_i$; 
for $i\in {\cal{V}}_l$, the node feature vector $v_i$ only includes the embedding feature $v^l_i$. 
For each edge $({i\rightarrow j})\in {\cal{E}}$ with $i,j\in {\cal{V}}$, the edge feature vector is denoted as $e_{ij}$. Functions $\phi$ and $\rho$ denote the update embedding function and aggregate function respectively.
As shown in Fig.~\ref{comm} and detailed in Table \ref{Edge-features}, the overall GNN-based communication sub-module consists of three steps. 

%\vspace{-0.2cm}
\noindent\textbf{Step 1) Intra-group aggregation.} In each group, the low-level agents embed their local information and send it to the associated high-level agent $j\in {\cal{V}}_h$; the high-level agents aggregate the information from all associated low-level agents and obtain the cluster perception;
% updates its central role feature;

%\vspace{-0.2cm}
\noindent\textbf{Step 2) Inter-group sharing.} The high-level agent communicates with the other high-level agent with cluster perception. This further aggregates all received other high-level messages to obtain the global perception;

%\vspace{-0.2cm}
\noindent\textbf{Step 3) Intra-group sharing.} Each of the high-level agents communicates all its features with the associated low-level agents while the low-level agents aggregate the received information from high-level agents. The embedding feature of both high-level and low-level agents are then updated.
% \begin{itemize}
% \item \textbf{Step 1: Intra-group aggregation.} In each group, the normal agent embeds their local information and sends it to the associated central agent $j\in {\cal{V}}_c$; the central agent aggregates the information from all associated normal agents and updates its central role feature;
% \item \textbf{Step 2: Inter-group sharing.} The central agent communicates the information with the other central agent with cluster information, further aggregates the received  and indicates the global perception;
% \item \textbf{Step 3: Intra-group sharing.} The central agent communicates all its feature with the associated normal agents while the normal agent aggregates the received information from central agents. Both the embedding feature of central and normal agents will be updated.
% \end{itemize}

\begin{table*}[tb!]
\centering
\caption{The proposed GNN-based communication architecture with three steps.}
\resizebox{\textwidth}{!}{\begin{tabular}{ c|c|c|c }
\hline
Type & Edge $(i\rightarrow j)\in {\cal{E}}$ & Edge Update Scheme & Node Update Scheme\\
\hline
Step 1: intra-group aggregation & $i\in {\cal{V}}_l $, $j\in {\cal{V}}_h$ & $e_{ij} = \phi (v_{i}^l)$, $\bar{e}_{j} = \rho (\left\{ e_{ij} \right\}_{(i\rightarrow j)\in {\cal{E}}})$ & $v_{j}^h = \phi ( \bar{e}_{j}, v_j^l)$\\
\hline
Step 2: inter-group sharing & $i\in {\cal{V}}_h$, $j\in {\cal{V}}_h$ & $e_{ij} = \phi (v_{i}^h,v_{i}^l)$, $\bar{e}_{j} = \rho (\left\{ e_{ij} \right\}_{(i\rightarrow j)\in {\cal{E}}})$ & $v_{j}^g = \phi ( \bar{e}_{j}, v_j^l)$ \\
\hline
Step 3: intra-group sharing & $i\in {\cal{V}}_h$, $j\in {\cal{V}}_l \cup {\cal{V}}_h$ & $e_{ij} = \phi (v_{i}^g,v_{i}^h,v_{i}^l)$, $\bar{e}_{j} = \rho (\left\{ e_{ij} \right\}_{(i\rightarrow j)\in {\cal{E}}})$ & $v_{i}^l = \phi ( \bar{e}_{i}, v_i^l)$, $v_{j}^l = \phi ( \bar{e}_{j}, v_j^l)$ \\
\hline
\end{tabular}}\label{Edge-features}
\end{table*}

%\textbf{Local Info Handler:} ordinary agents generate and send local information to central agents, then central agents try to obtain cluster perception from received messages.
%\begin{equation}
%	e_k^{nc}=\phi^{enc}(v^n_{s_k^{nc}})
%\end{equation}
%\begin{equation}
%	\overline { e }^{nc} _ { i } = \rho ^ { e^{nc} \rightarrow v } \left( E _ { i } ^ { nc } \right)
%\end{equation}
%\begin{equation}
%	 v _ { i } ^ { c } = \phi ^ { vc } \left(    e _ { i } ^ { \prime } ,  v _ { i }^n \right)
%\end{equation}

%\textbf{Cluster Info Handler:} central agents communicate with each other with cluster information, then aggregate received messages to get the global perception.
%\begin{equation}
%	e_k^{cc}=\phi^{ecc}(v^c_{s_k^{cc}},v^n_{s_k^{cc}})	%
%\end{equation}
%\begin{equation}
%	\overline { e }^{cc} _ { i } = \rho ^ { e^{cc} \rightarrow v } \left( E _ { i } ^ { cc } \right)
%\end{equation}
%\begin{equation}
%	v _ { i } ^ { g } = \phi ^ { vg} \left(   \overline{e }_ { i }^{cn} ,  v _ { i }^n \right)
%\end{equation}

%\textbf{Order Handler: }central agents order to ordinary agents to guide cooperation inner group. After that normal agents will update its perception based on received messages.
%\begin{equation}
%	e_k^{cn}=\phi^{ecn}(v^g_{s_k^{cn}},v^c_{s_k^{cn}},v^n_{s_k^{cn}}, e_k^{nc})
%\end{equation}
%\begin{equation}
%	v _ { i } ^ { n } = \phi ^ { vn} \left(   \overline{ e} _ { i } ^ { nc } ,  v _ { i }^n \right)
%\end{equation}
The GNN-based communication sub-module is modeled as a GNN ($f_{\theta^{gnn}}$) with parameter $\theta^{gnn}$, while the following $Q$-Net of agent $i$ ($Q^{i}_{\theta^Q}$) is parameterized by shared parameter $\theta^{Q}$.
The gradient can be back-propagated from $Q$-Net to the graph neural network. As a result, the overall loss of communication-based policy module is as follows:
\begin{equation} \label{eq:loss-gnn}
\ell (\theta^{Q},\theta^{gnn}) = \mathbb { E } _ {\mathbf{o}, \mathbf{a} , \mathbf{r} , \tilde{\mathbf{o}} } \left[ \sum_{i=1}^n \left( Q^{i}_{\theta^Q} ( f_{\theta^{gnn}} (\mathbf{o}), a_i ) - y_i \right)^{2} \right],
\end{equation}
where $y_i = r_i + \gamma \max_{ \tilde{a}_i } Q^{i}_{\theta^Q} (f_{\theta^{gnn}} (\tilde{\mathbf{o}}), \tilde{a}_i )$, and $r_i$ is the reward for agent $i$. 
Soft updating scheme is used:
\begin{equation} \label{eq:target-network}
\begin{split}
\theta^{\tilde{Q}} =\tau \theta^{Q}+(1-\tau) \theta^{\tilde{Q}}, \\
\theta^{\tilde{gnn}} =\tau \theta^{gnn}+(1-\tau) \theta^{\tilde{gnn}}. 
\end{split}
\end{equation}
% \vspace{-0.3cm}

The whole LSC is depicted in Algorithm \ref{alg:LSC}. The CBRP function automatically and distributively establishes the structured communication network based on the learned importance weights. HCOMM denotes the communication-based policy module, which outputs the $Q$-values based on the GNN-based communication messages. The details of CBRP and HCOMM can be found in the Appendix.
\begin{algorithm}[tb!]
\caption{{\bf{LSC}}: Learning Structured Communication}
\label{alg:LSC}
\begin{algorithmic}[1]
\State \textbf{Initialization:} weight generator parameter: $\theta^w$, $Q$-net: $\theta^{Q}$, GNN: $\theta^{gnn}$, target $Q$-net: $\theta^{\tilde{Q}}$, target GNN: $\theta^{\tilde{GNN}}$ replay buffer ${\cal{R}}=\varnothing $, cluster radius $d$, the number of agents $n$;
\For{${\hbox{Episode}} = 1, \cdots, M$}
\State Reset $t=0$, global state $s^t$ and observation $o^t_i$ for each agent $i$, low-level agents set ${\cal{V}}^t_l=\{\text{all agents}\}$ and high-level agents set ${\cal{V}}_h^t=\varnothing$;
    \For{$t=1,\cdots,T$ and $s_t \neq$ terminal}
        \For{each agent $i$}
            %\State $q_{\text{w}_i^t}=Q_{\theta^w}(o_i^t)$
            \State With probability $\epsilon$ pick a random action $w_i^t$ else $w^t_i=$\;$\arg\max_{\left\{ w_i \right\}} Q_{\theta^w}(o_i^t)$;
        \EndFor
        \State Get current position $\text{POSs}_i^{t}$ of each agent $i$;
        \State $({\cal{V}}_l^t, {\cal{V}}_h^t, {\cal{E}})$\;=\;CBRP($({\cal{V}}_l^{t-1}, {\cal{V}}_h^{t-1})$,$\{w_{1}^t, \cdots, w_{n}^t\}$,\;\\ \hspace{0.2in} $\{\text{POSs}_1^{t}, \cdots, \text{POSs}_n^{t}\}$,\;$d$);
        \State $\{q^t_{1}, \cdots, q^t_{n}\}$\;=\;HCOMM(${\cal{V}}_l^t$,\;${\cal{V}}_h^t$,\;${\cal{E}}$);
        \For{each agent $i$}
            \State With probability $\epsilon$ pick a random action $a_i^t$ else choose action with the largest value in $q^t_i$;
        \EndFor
        \State Execute global actions and get global reward $r^t$ , next state $s^{t+1}$ , next observation $o^{t+1}$;
        \State Get updated position $\text{POSs}^{t+1}_i$ for each agent $i$;
        \State Store $(s^t, o^t, \{\text{POSs}_1^{t}, \cdots, \text{POSs}_n^{t}\}, a^t, r^t, o^{t+1},$ \\ \hspace{0.2in} $\{\text{POSs}_1^{t+1}, \cdots, \text{POSs}_n^{t+1}\}, s^{t+1})$ to $\cal{R}$;
    \EndFor
    \For{$k = 1, \cdots, K$}
        \State Sample a random mini-batch transitions from $\cal{R}$;
        \State Update weight generator $\theta^w$ by Eq.~(\ref{eq:loss-structure});
        \State Update communication based policy module $(\theta^Q,\theta^{gnn})$ by minimizing Eq.~(\ref{eq:loss-gnn});
        \State Update the target networks through Eq.~(\ref{eq:target-network}).
    \EndFor
\EndFor
\end{algorithmic}
{\footnotesize Refer to Appendix for details of HCOMM and CBRP.}
\end{algorithm}

\subsection{Communication Efficiency Analysis}
We discuss the communication efficiency\footnote{Communication efficiency varies by different communication mechanisms. Here our analysis is under the peer to peer mode.} of our LSC from three aspects: the number of messages exchanged ($N_{msg}$) among agents; the number of communication steps before acts ($N_{step}$); the max communication bandwidth required for an agent ($N_{b{\text{-}}r}$).
%The required bandwidth in practical scenarios is %related to the bits of messages, the number of %messages and many other factors, only the number of %messages is close related to communication %topologies. 
Here the bandwidth is measured by the number of messages sent/received by the agent in each episode.
%, because this number is proportional to the usage of bandwitdth. 

%{\color{red} THChang: The last setentence is not understandable to me.}
%{\color{blue} Jin: The required bandwidth is not easy to measure, so we say: the amount of communication can be counted by the message number, then if the time for each step in episode is fixed, the bandwidth = amount / time.}
%\jar{revision: The required bandwidth is related to many practical factors, while for MARL communication, the most influential factor of bandwidth is the number of sent/received messages at each step. Thus we choose it as the analytical measures for required bandwidth.}

Table \ref{tb:efficiency} compares the communication efficiency of different communication structures. 
In fully-connected (FC) structures where each agent communicates with all the others, the message exchanging complexity is ${\cal{O}}(n^2)$, and the max bandwidth for an agent is ${\cal{O}}(n)$. In the star structure, agents only need to communicate with the central agent, and thus the message exchanging complexity decrease to ${\cal{O}}(n)$. 
For the tree structure and neighboring structure, agents only need to communicate with neighbors. 
We denote the number of groups as $k$ and the maximum number of agents in a group as $b$. 
The tree structure and neighboring structure need ${\cal{O}}(kb^2)$ message exchanging complexity. 
The tree structure lets groups communicate sequentially, thus need ${\cal{O}}(b)$ communication steps. %{\color{red} ${\cal{O}}(b)$ or ${\cal{O}}(d)$ ? The table shows ${\cal{O}}(d)$ for the tree.}
Our hierarchical communication structure only needs low-level agents to communicate with the high-level agents, and high-level agents need to communicate with each other. 
Thus the message exchanging complexity is ${\cal{O}}(kb+k^2)$. 
Since $k$ and $b$ increase mildly with $n$, the proposed dynamic communication topology in LSC is suitable for large-scale MARL.
\begin{table}[tb!]
%\footnotesize
\caption{Structure comparison for communication efficiency.}
\centering
\begin{tabular}{ |c|c|c|c|c|c| } 
\hline
  & \makecell[c]{FC} & Star & Tree & \makecell[c]{Neighbor} & Hierarchical   \\
\hline
$N_{msg}$ & ${\cal{O}}(n^2)$ & ${\cal{O}}(n)$ & ${\cal{O}}(kb^2)$ & ${\cal{O}}(kb^2)$ & ${\cal{O}}(k^2 + kb)$ \\ 
\hline
$N_{step}$ & ${\cal{O}}(1)$ & ${\cal{O}}(1)$ & ${\cal{O}}(b)$ & ${\cal{O}}(1)$ & ${\cal{O}}(1)$ \\ 
\hline
$N_{b{\text{-}}r}$ & ${\cal{O}}(n)$ & ${\cal{O}}(n)$ & ${\cal{O}}(b)$ & ${\cal{O}}(b)$ & ${\cal{O}}(b+k)$ \\ 
\hline
\end{tabular}\label{tb:efficiency}
% \vspace{-0.5cm}
\end{table}

\section{Experiments}
We choose MAgent\footnote{https://github.com/geek-ai/MAgent} and MPE\footnote{https://github.com/openai/multiagent-particle-envs} as our experiment platforms.
% MAgent\footnote{https://github.com/geek-ai/MAgent} and MPE\footnote{https://github.com/openai/multiagent-particle-envs} are adopted as the experiment platforms, due to their support for the large-scale MARL setting. 
Specifically, two scenarios are tested: i) battle and ii) cooperative spread. 
For compared baselines, first, we choose IDQN~\cite{Tampuu2017}, which can be regarded as a degenerated version of our LSC with disabled communication. 
We also replace the adaptive hierarchical grouping component in our LSC with the star and neighboring topology for communication, as in CommNet~\cite{Sukhbaatar2016LearningMC} and DGN~\cite{Jiang2020Graph}, respectively. 
We call these two variants LSC-star, LSC-nbor, and they serve as two benchmarks.
They are all trained using Q-Learning, and the parameters are shared among agents. 
We refrain from directly comparing the proposed LSC with CommNet and DGN because they involve quite different techniques, and it is difficult to have a meaningful comparison. 
For a fair comparison, the basic hyperparameters of all methods under test are set the same. 
We adopt the graph\_nets\cite{battaglia2018relational} to implement the GNNs.  
The details of settings and hyperparameters are in the Appendix.

% LSC and all the baselines are parameter-sharing and the basic hyperparameters are kept to the same. For reproducibility, the detailed experimental settings and hyperparameters are provided in Appendix.
% We compare LSC with state-of-the-art MARL methods in two large-scale battle environments, i.e., the grid world platform MAgent~\cite{Zheng2017MAgentAM} and StarCraft2~\cite{samvelyan19smac}, to evaluate their performances from aspects of both network structure and communication.

%\vspace{-10pt}

\subsection{Task I: Large-scale Battle Game in MAgent} 
%\vspace{-0.2cm}
In this scenario (self-interested cooperative scenario), $n$ agents move and fight against $\ell$ enemies. The enemies have higher speed, higher attack power, and better stamina. Thus, agents need to form high-quality cooperation to wipe out enemies. The enemies are controlled by an IDQN~\cite{Tampuu2017} pretrained policy. To evaluate different methods, besides the learning curve, we choose some quantitative evaluation criteria of the battle game, like `Mean-reward' (average per-step reward of all agents), `$N_k$' (average number of kills per episode), `$N_d$' (average number of deaths per episode) and `$r_{kd}$' (kill to death ratio $N_k/N_d$).
%properties of the game (mean-reward, kills, death and kill-death ratio) as the basic metrics. Mean reward denote the average reward an agent achieve at a single step. Kills($N_k$) and death($N_d$) denotes the average numbers of kills and death agent in an episode. We also calculate the kill-death ratio to show the performance of cooperation.

%Moreover we make the visualization of strategies to find out why the performance are different.

\begin{figure*}[tb!]
	\centering
	\subfigure[Structure comparison]{
	\includegraphics[width=0.45\textwidth]{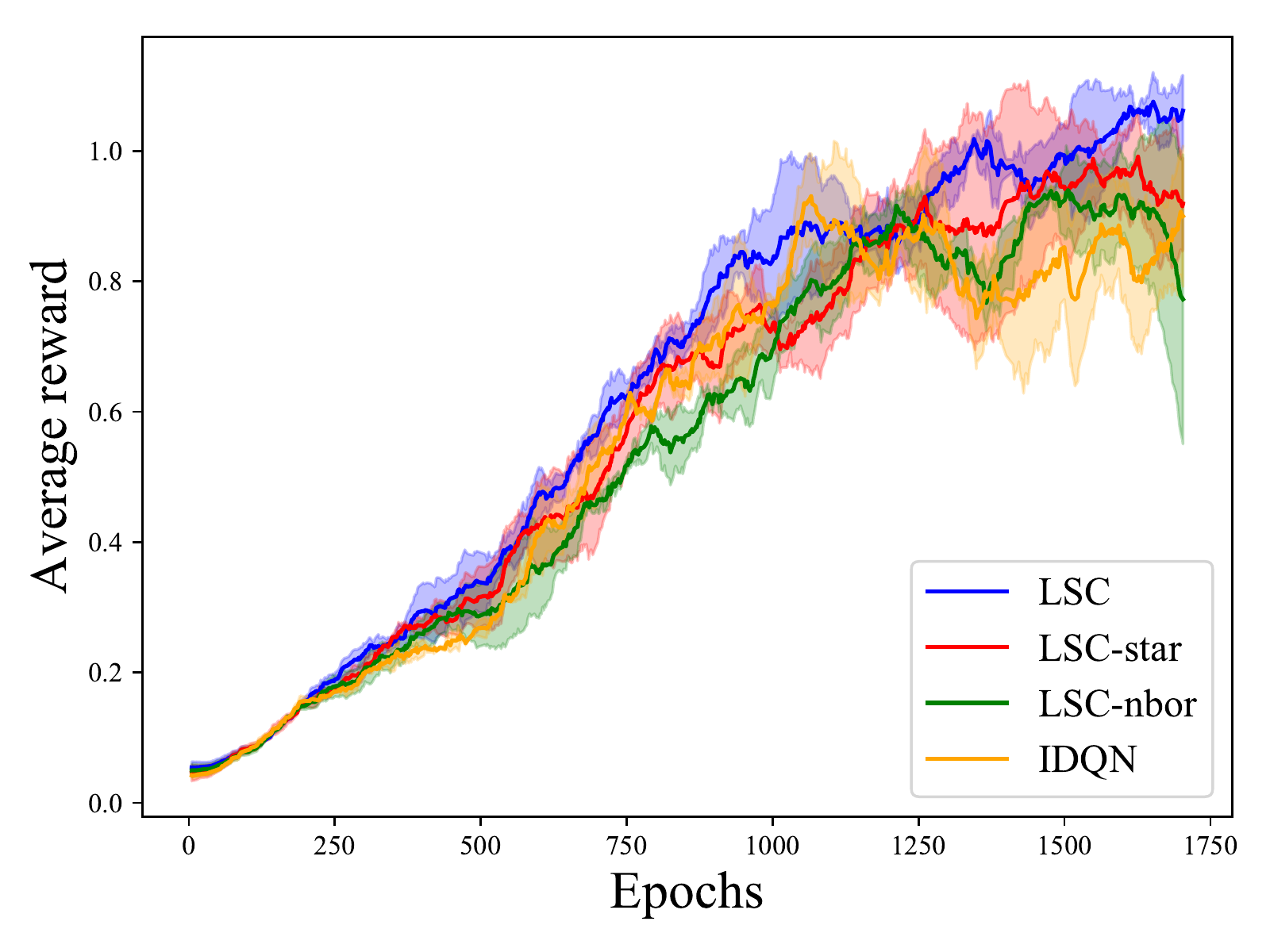}
	\label{fig:averagereward}
	}	
	\subfigure[Weight generator comparison]{
	\includegraphics[width=0.45\textwidth]{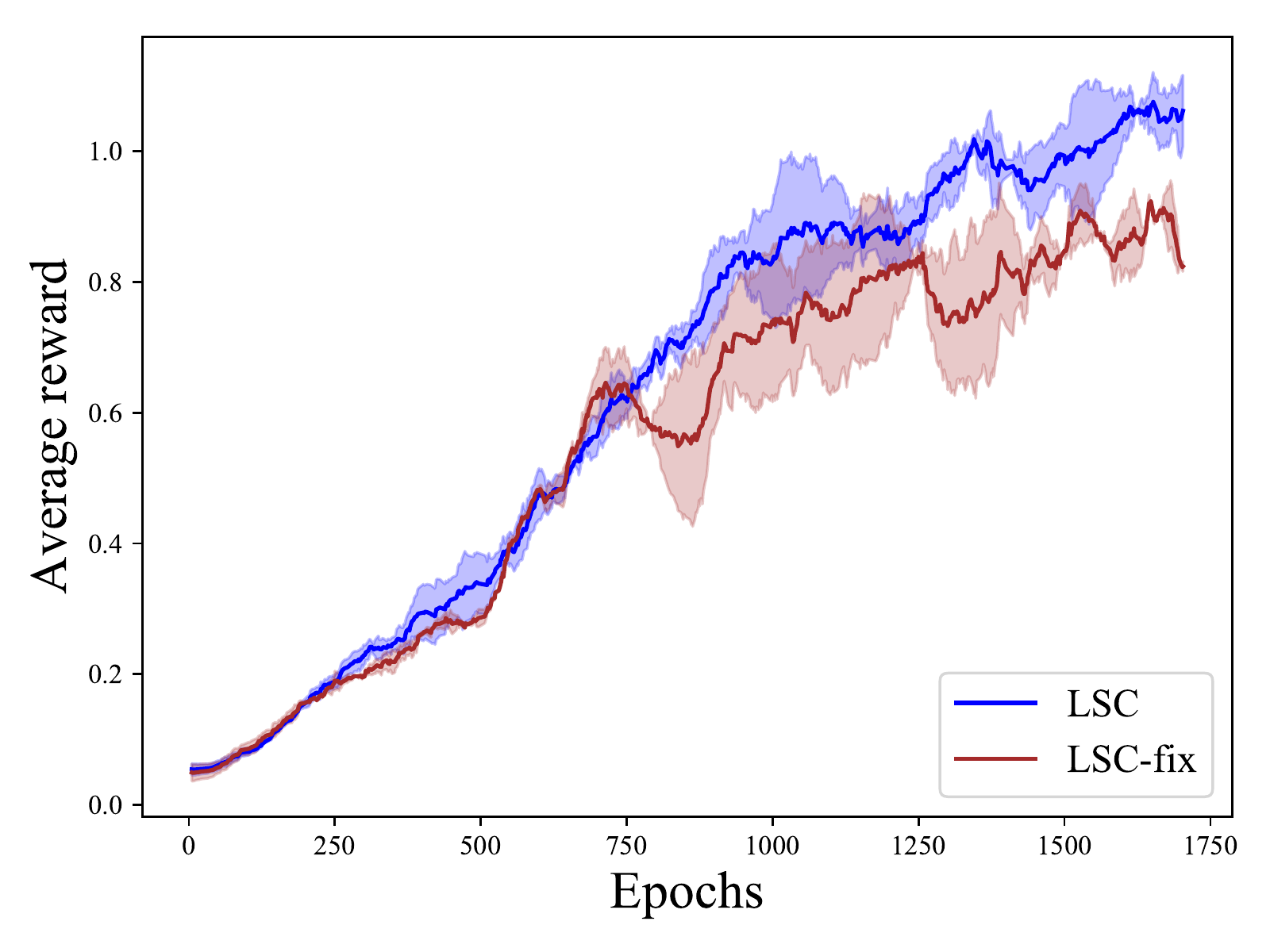}
	\label{fig:wg}
	}	
	\subfigure[LSC and LSC-star-gate]{
	\includegraphics[width=0.45\textwidth]{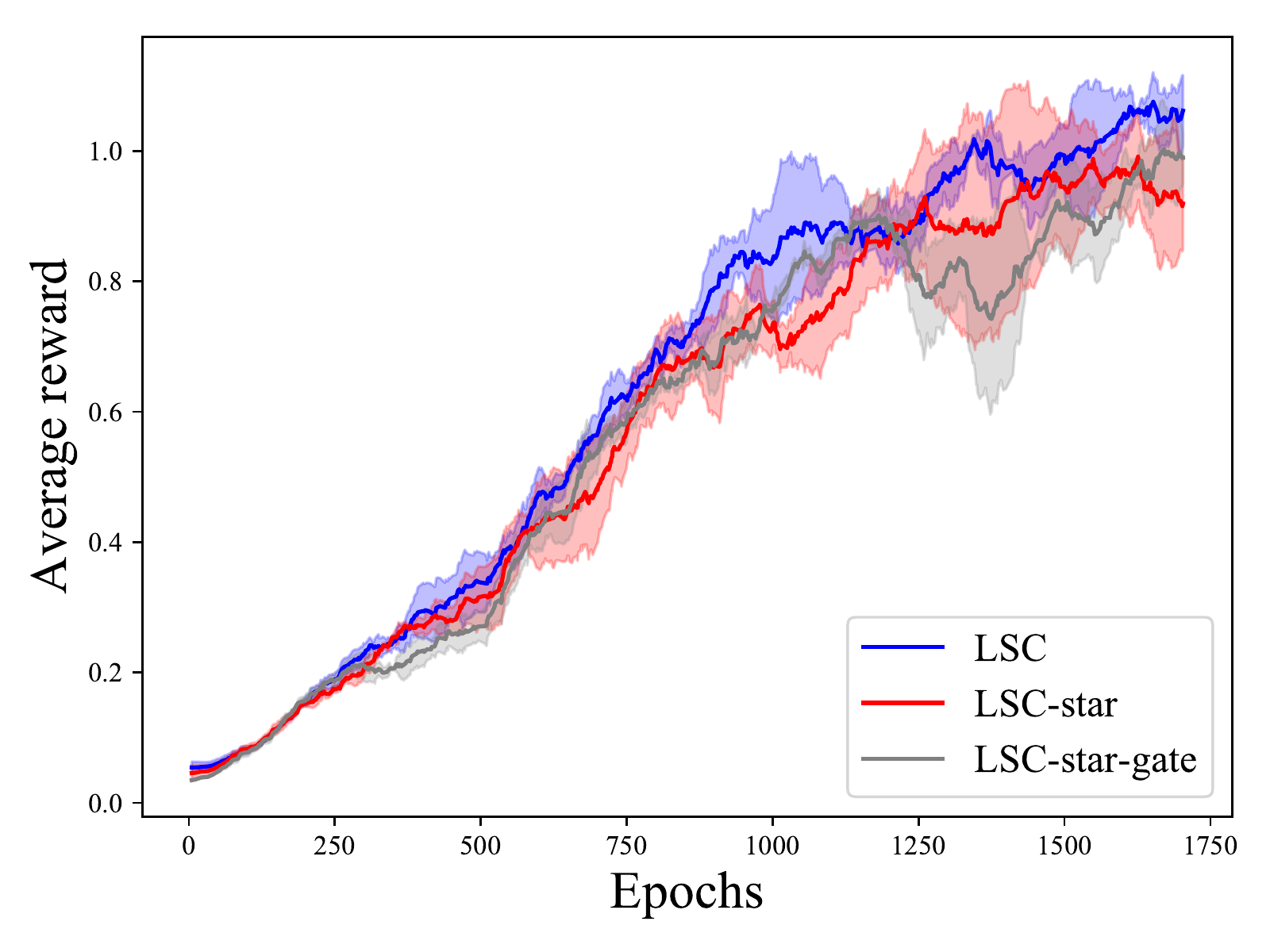}
	\label{fig:add-com}
	}
% 	\vspace{-8pt}
	\caption{Comparisons of learning curves in Battle game.}
\end{figure*}

We first compare the performance of different communication structures.
All the models are trained with $n=64$ and $\ell=64$ for $1750$ episodes. Fig.~\ref{fig:averagereward} shows the learning curve. The solid and shadow denote the mean and variance, respectively. As seen, LSC performs better in terms of the converged mean reward. We further test the learned models for $50$ rounds, and the results are shown in Table~\ref{tb:battle-eval}. One can observe that LSC achieves a higher mean reward and a larger kill-death ratio. IDQN yields the lowest score due to a lack of communication schemes. Since LSC 
has additional intra-group communication when compared to LSC-nbor and extra inter-group communication between high-level agents when compared to LSC-star, the performance improvement of LSC over the two benchmarks well demonstrates the benefits brought by a hierarchical structure, intra-inter group communications.

% Although LSC uses original reward feedback to establish the weight generator, the weight generator can use a predefined rule or add additional feedback (i.e. latency or the resources consumption) to the generator learning. 
To investigate the impact of the learned importance weight generator, we compare LSC with a counterpart with a basic fixed weight generator (all agents are set to the same communication weight) in the battle scenario. In Fig.~\ref{fig:wg}, our learned weight generator significantly outperforms the fixed one (LSC-fix). The result indicates that the structured communication module and communication-based policy module in LSC are strongly connected. We also compare LSC with LSC-star-gate (which incorporates communication gates in the star structure like IC3). In Fig.~\ref{fig:add-com}, one can see that LSC outperforms LSC-star-gate, and therefore the communication gate is barely beneficial.
%Although adding the communication gate can help agents learn to mute, the central agents in the star network still need to handle all the messages.
%, without a hierarchical structure.

%It is interesting that LSC-star achieve higher mean-reward than LSC-nbor while got lower kill-death ratio. 
%Because agents are rewarded by attacking an enemy and enemies can recover with times, agents controlled by LSC-star are thus less like to kill enemies than LSC-nbor.

\begin{figure}[tb!]
	\centering
	\subfigure[Initial state]{ 
		\includegraphics[width=0.22\textwidth]{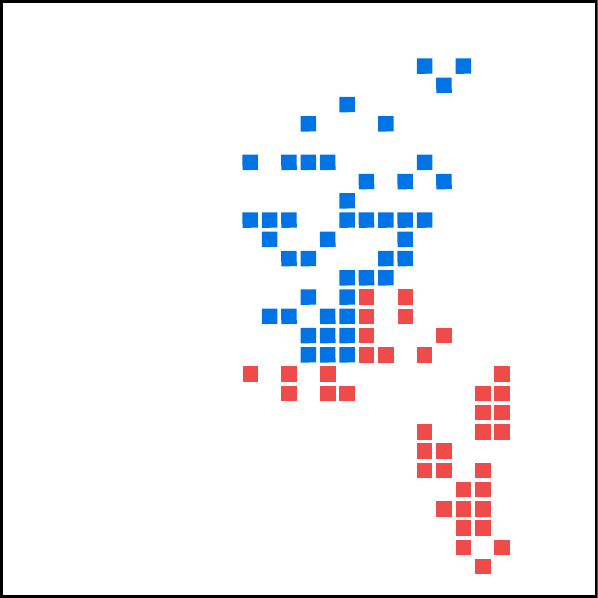}
		\label{fig:init}
	}
	\subfigure[Encircle]{ 
		\centering
		\includegraphics[width=0.22\textwidth]{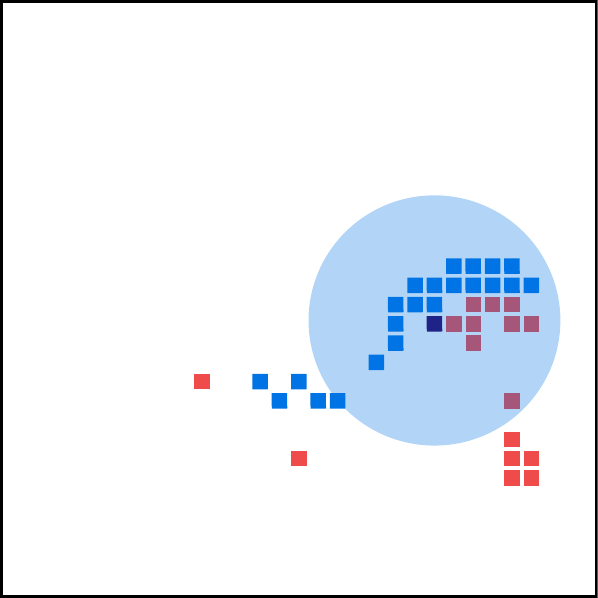}
		\label{fig:intra-group-1}
	}
	\subfigure[Fire focusing]{ 
		\centering
		\includegraphics[width=0.22\textwidth]{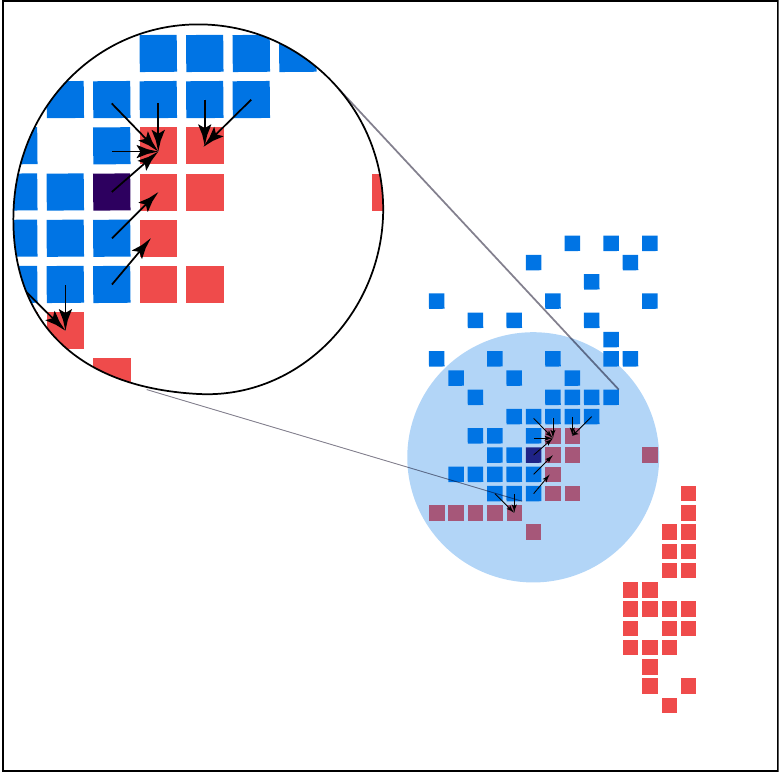}
		\label{fig:intra-group-2}
	}\\
	%-- Early state
	\subfigure[LSC]{ %Early state
		\includegraphics[width=0.22\textwidth]{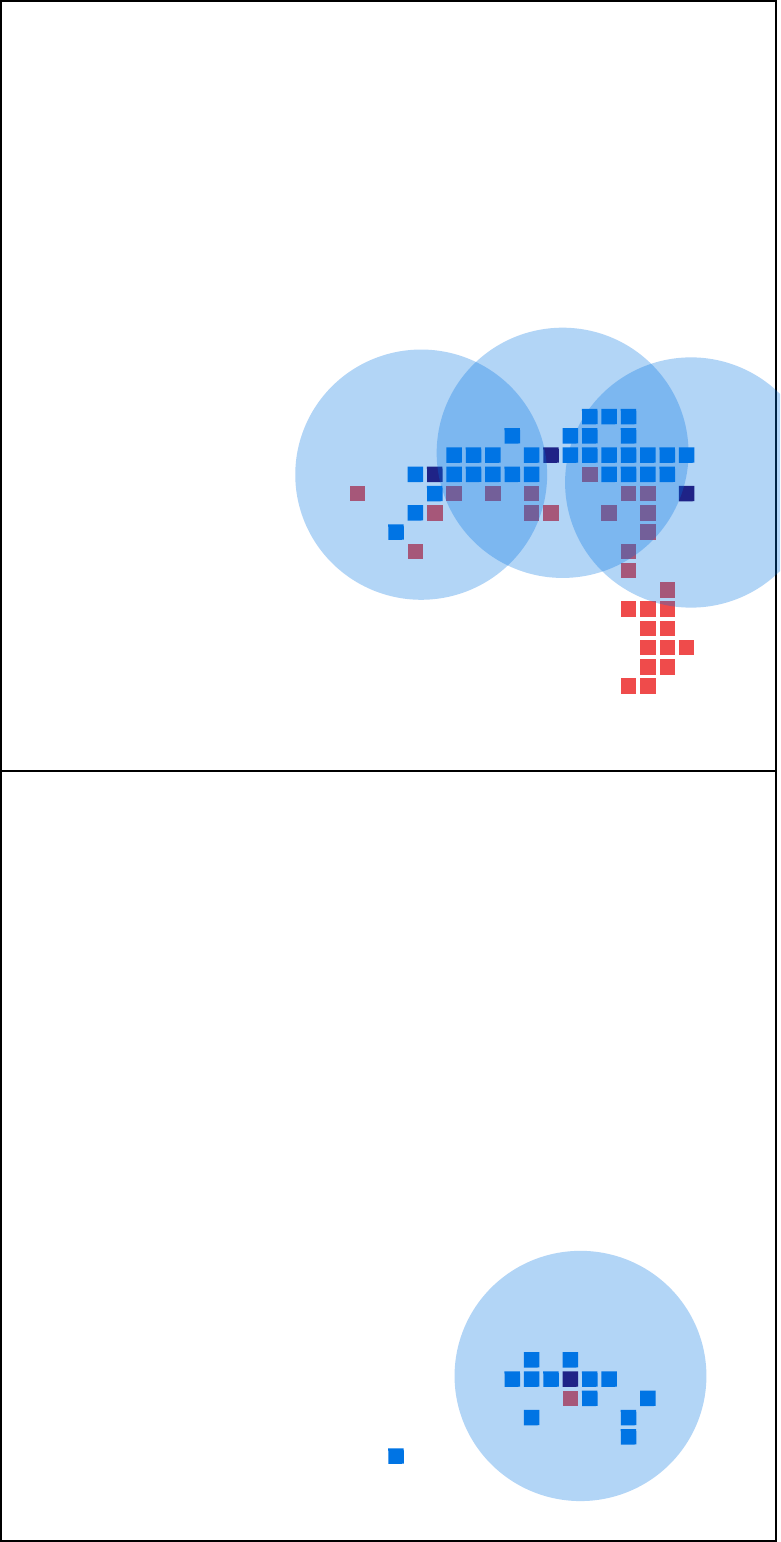} 
        \label{fig:LSC-ES}
	}
	\subfigure[IDQN]{ %Early state
		\includegraphics[width=0.22\textwidth]{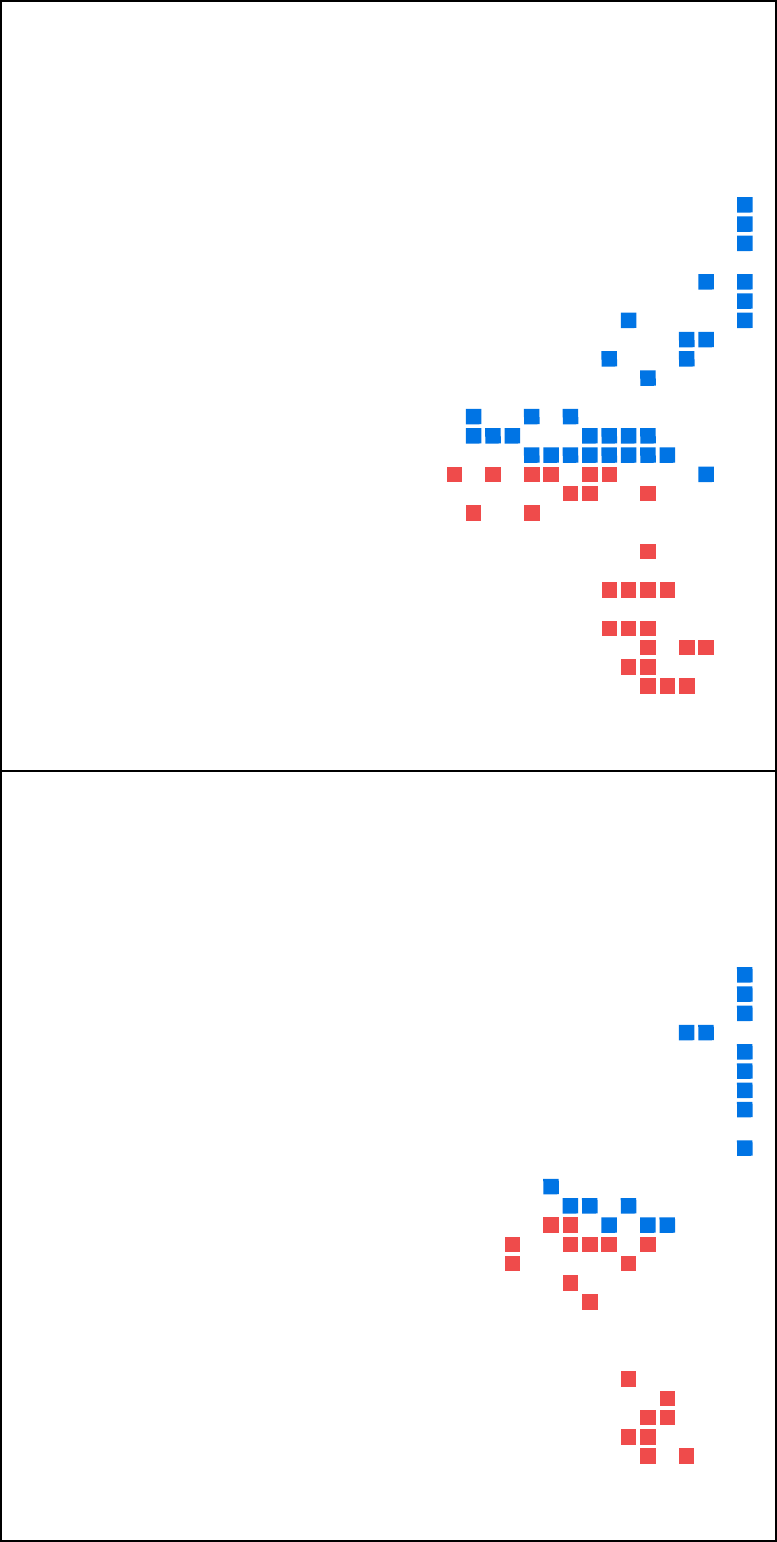}
		\label{fig:idqn-ES}
	}
	\subfigure[Star]{ %Early state
		\centering
		\includegraphics[width=0.22\textwidth]{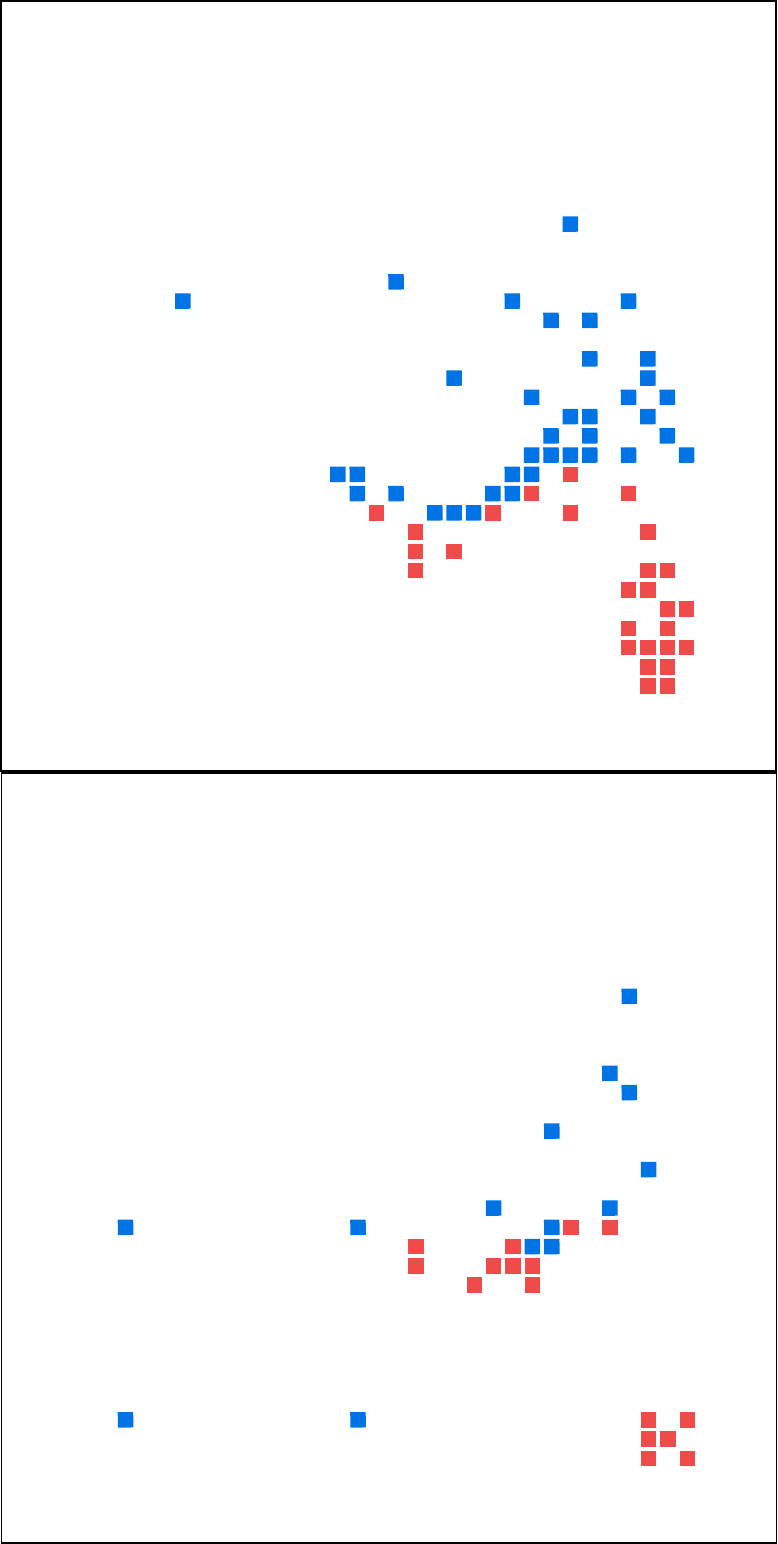} 
		\label{fig:comm-ES}
	}
%     \subfigure[IC3]{ %Early state
% 		\centering
% 		\includegraphics[width=0.14\textwidth]{./figures/ic3}
% 		\label{fig:ic3-ES}
% 	}
%     \subfigure[ATOC]{ %Early state
% 		\centering
% 		\includegraphics[width=0.14\textwidth]{./figures/atoc}
% 		\label{fig:atoc-ES}
% 	}
	\subfigure[Neighbor]{
	    \centering
	    \includegraphics[width = 0.22\textwidth]{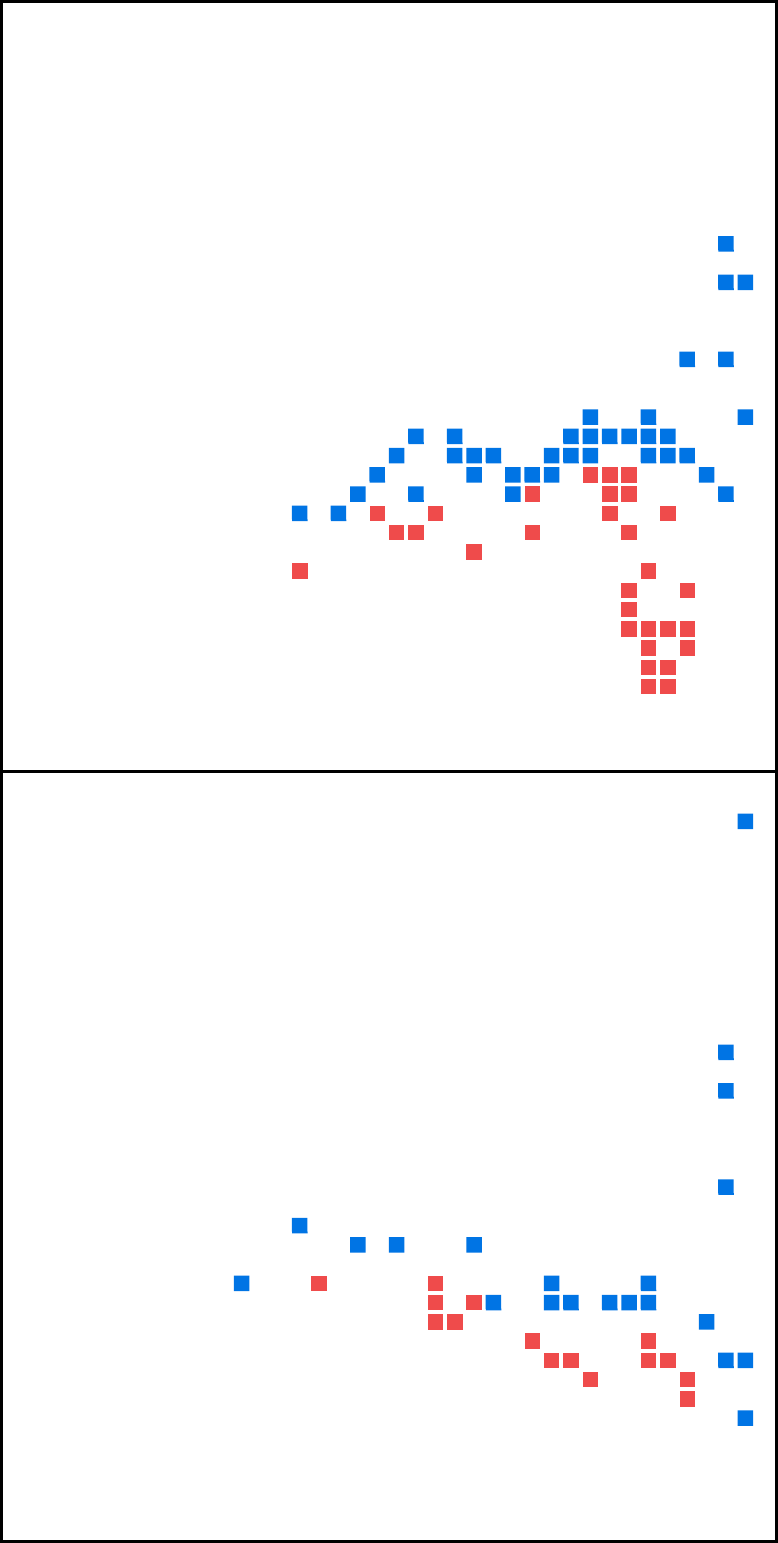}
	    \label{fig:neigh-ES}
	}
% 	\subfigure[MFQ]{ %Early state
% 		\centering
% 		\includegraphics[width=0.14\textwidth]{./figures/mfq}
% 		\label{fig:mfq-ES}
% 	}
% 	\vspace{-10pt}
	\caption{Behavior illustration. The first row shows two typical behavior by LSC. In the second row, the top and bottom plot denote the early state and the near to final battle state, respectively.}%\vspace{-0.5cm}
    \label{fig:behavior}
\end{figure}

Fig.~\ref{fig:behavior} is presented to better understand the strategies learned by these algorithms. 
Specifically, LSC learns the encircle and fire focusing strategies as shown in Fig.~\ref{fig:intra-group-1} and Fig.~\ref{fig:intra-group-2}. 
We find that the baselines can hardly handle the situation when some agents are far away from enemies. 
For example, the situation in Fig.~\ref{fig:init}, the agents in the top right can not know where to attack without communication. 
We let every learned model run from the initial state, and find only LSC learns to form an inter-group encircle and wipe out the enemies (see Fig.~\ref{fig:LSC-ES}). IDQN tends to cooperate within the visual range. Thus the agents that find no enemy would get close to the wall to defend attacks. 
Local cooperation leads to failed results, as shown in Fig.~\ref{fig:idqn-ES}. Star and neighbor structures help agents form global cooperation. 
However, the agents far away from the majority have difficulty in comprehending the global information. 
In Fig.~\ref{fig:comm-ES} and Fig.~\ref{fig:neigh-ES}, such remote agents choose the spread out and exploration strategy, thus fail to wipe out enemies. 
To sum up, in Fig.~\ref{fig:LSC-ES}, agents controlled by LSC form a global encircle strategy by communication in both intra-group and inter-group, which can wipe out enemies more efficiently.

\begin{figure}[tb!]
	\centering
	\subfigure[Before CBRP]{ 
		\includegraphics[width=0.25\textwidth]{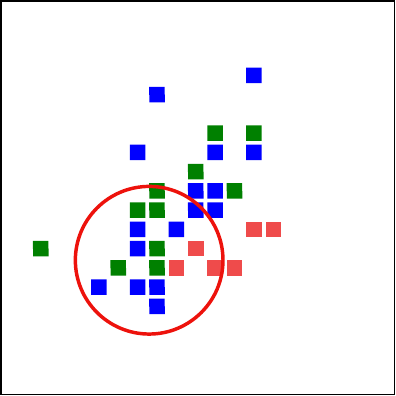}
		\label{fig:weight}
	}
	\subfigure[After CBRP]{
		\centering
		\includegraphics[width=0.25\textwidth]{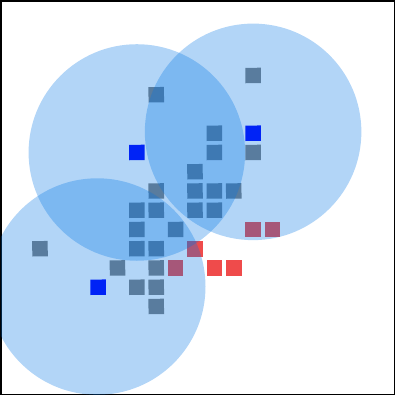}
		\label{fig:ch}
	}
% 	\vspace{-10pt}
	\caption{Weight visualization of dynamic communication structure: `Before CBRP' and `After CBRP' stages.
	Blue, green and grey nodes denote agents with weight $2$, $1$ and $0$ (enemies in red).}
    \label{fig:weight-vis}
    %\vspace{-0.4cm}
\end{figure}

Fig.\ref{fig:weight-vis} presents two weight-visualization graphs to show how the CBRP sub-module works. 
\ref{fig:weight} visualizes the agent weights of an intermediate stage during the testing procedure for a $64\times 64$ battle. 
Red agents denote the enemies. 
As discussed in Section \ref{subsec:CSNM}, only three kinds of discrete weights can be obtained for each agent, i.e., $\{0,1,2\}$. 
The blue agents denote the agents have weight 2, and the green ones denote the agents have weight 1, while there is no agent with weight $0$ in this stage. 
Without the CBRP sub-module, all the blue agents with weight $2$ will be elected as high-level agents. 
This leads to an almost dense high-level agent structure, for instance, the red circle area.
\ref{fig:ch} visualize the agent weights after implementing CBRP method.
Only three agents are set to be high-level agents (weight $2$), while all others are set to be low-level agents (weight $0$).

\begin{table}[tb!]%{0.5\textwidth}
%\footnotesize
\centering
%\vspace{-15pt}
\caption{Performance comparisons of $64$ vs. $64$ case in $50$ testing trials on Battle game. The bold denotes the best result in each row.}
% \vspace{10pt}
\begin{tabular}{ |c|c|c|c|c|c|c| } 
\hline
Criteria / Method   & LSC  & LSC-star & LSC-nbor & IDQN\\
\hline
Mean-reward    & $\bf{1.13}$ & $1.04$ & $0.94$  & $0.89$\\ 
\hline
$N_{k}$     & $\bf{62.9}$  & $61.36$ & $62.4$ & $59.3$\\ 
\hline
$N_{d}$    & $\bf{39.4}$  & $48.52$ & $43.6$ & $55.8$\\ 
\hline
$r_{kd}$   & $\bf{1.60}$  & $1.21$ & $1.43$ & $1.06$\\
\hline
\end{tabular}
% \vspace{-30}
\label{tb:battle-eval}
\end{table}

\subsection{Task II: Cooperative Spread in MPE}
\begin{figure}[tb!]
    %\vspace{-10pt}
	\centering
	\includegraphics[width=0.45\textwidth]{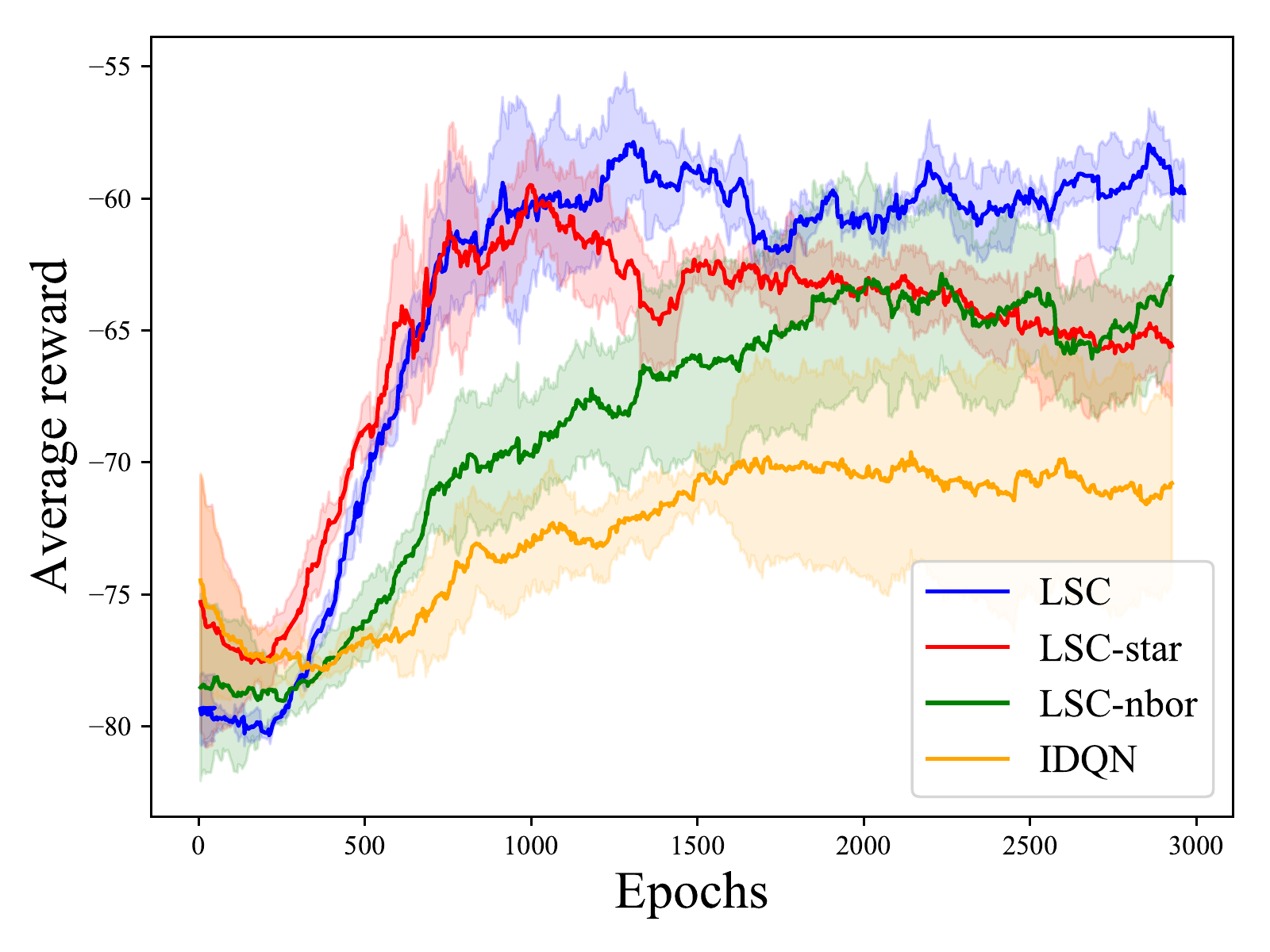}
% 	\vspace{-15pt}
	\caption{Learning curves on Cooperative Spread scenario in MPE.}
	\label{fig:spread-lc}
\end{figure}

We design a new scenario: cooperative spread based on MPE~\cite{lowe2017multi},  to test the performance of LSC in the fully cooperative scenario. There are $12$ agents and $4$ landmarks in this environment. Every landmark needs to be reached by three agents. However, when there are more than three agents reaching the landmark, the landmark would be overloaded and penalizes the agents. We train LSC and other baselines with $3000$ episodes, whose learning curves are shown in Fig.~\ref{fig:spread-lc}. We take $50$ test rounds on the obtained models, and Table~\ref{tb:spread-eval} contains some evaluation criteria in the testing procedure, for instance `$N_s$' (number of successive reaching of three agents to one landmark), `$N_o$' (number of successive reaching of more than three agents to one landmark) and `Mean-reward' (average per-episode reward of all agents).

Fig.~\ref{fig:spread-lc} and Table \ref{tb:spread-eval} show LSC outperforms in training and testing. The high-level agents (in star or hierarchical topology) can help speed up the learning process by making an agreement of global information. Thus LSC and star topology learn faster than the other two. LSC converges to a higher reward than baselines. Table \ref{tb:spread-eval} shows IDQN's strategy is passive. Agents avoid overload while taking less chance to reach the landmark. Star and neighboring structures take more aggressive strategies, and the star structure cannot achieve fine-grained information from neighbors, thus lead to more overloads. The neighboring structure makes some agents disconnected to others, thus cannot achieve global communication and lead to less chance for success. LSC achieves the highest reward during testing, with reasonable overload.

% \jar{take the zero shot generalizetion}
\begin{table}[tb!]%{0.5\textwidth}
%\footnotesize
\centering
% \vspace{-10pt}
\caption{Performance comparisons of $12$ agents case in $50$ testing trials). 
The bold stands for the best result in each row.}
% \vspace{10pt}
\scalebox{0.9}{\begin{tabular}{ |c|c|c|c|c| } 
\hline
Method    & LSC  & LSC-star & LSC-nbor &  IDQN \\
\hline
$N_s$ & $\bf{156}$  & $80$ & $61$ & $16$ \\ 
\hline
$N_o$ & $13$  & $16$ & $9$ & $\bf{2}$\\
\hline
Mean-reward & $\bf{-54.6}$  & $-68.2$ & $-69.1$ & $-78.1$\\
\hline
\end{tabular}
} 
% \vspace{-0.5cm}
\label{tb:spread-eval}
\end{table}

%\vspace{-0.3cm}
\section{Conclusion and Future Work}
% \vspace{-0.1cm}
In this paper, a novel learning structured communication (LSC) algorithm has been proposed for multi-agent reinforcement learning. The hierarchical structure is self-learned with a clustering-based routing protocol. The communication message representation is then naturally embedded and extracted via a graph neural network. Experiments on two scenarios demonstrate that our LSC can outperform existing learning-to-communicate algorithms with better communication efficiency, cooperation capability, and scalability. In the future, it is worthwhile to improve LSC by considering some practical constraints such as communication bandwidth and latency.
%\section{Acknowledgments}
%\clearpage

\small
\bibliographystyle{plain}
%\balance
\bibliography{example_paper}

\newpage
\appendix

\section{Appendix}

\subsection{Hyperparameters and Experimental settings}
% To enable reproducibility, we summarize the hyper-parameters as follow:
In MAgent battle, agents fight enemies in a $40\times 40$ grid world. Each agent in both sides has a $6\times6$ perception field and can attack its 8-adjacent grids. The speed, attack power, and health point for each agent are $1$, $1$, and $4$, which are increased to $2$, $2$, and $10$ for the enemy to increase the difficulty. 
The reward is $+5$ for successful attacking an enemy, $-2$ for being killed, and $-0.01$ for attacking a blank grid.

In the cooperative spread, 12 agents need to cooperate to reach every landmark with three agents. Each agent can get the relative position of other agents while only when the landmark in its receptive field(distance smaller than $0.4$) agent can get the landmark's relative position. The action space contains UP, DOWN, LEFT, RIGHT, and STAY. When three agents reach a landmark(distance smaller than 0.2), the landmark will reward to all the agents with $2$. However, if there are more than three agents reach a landmark, all the agents are penalized with $-10$. To help agents learn to reach landmarks, we add dense reward (landmarks give the negative-sum nearest three agents distance as a reward) like the typical spread setting.

To enable reproducibility, we summarize the hyperparameters for LSC and baselines at table \ref{tb:hyperparas}. The weight generator's hyper-parameters are all the same as IDQN, but the output layer sets to 3(the level of weights).

\begin{table}[htb!]%{0.5\textwidth}
\footnotesize
\centering
\caption{Hyperparamaters for LSC and baselines}
\begin{tabular}{c c c c c c c} 
\hline
Parameter    & LSC & IDQN & LSC-star & LSC-nbor  \\
\hline
discount($\gamma$) & \multicolumn{4}{c}{0.98} \\
batch size & \multicolumn{4}{c}{1/64}   \\
$\epsilon_{start}$     & \multicolumn{4}{c}{1.0}\\
$\epsilon_{end}$    &  \multicolumn{4}{c}{0.01}\\ 
\hline
optimizer & \multicolumn{4}{c}{Adam} \\
learning rate    & \multicolumn{4}{c}{$1e^{-4}$/$1e^-2$} \\
\hline
Conv layers & \multicolumn{4}{c}{2/---} \\
Aggregation & \multicolumn{4}{c}{seg sum} \\
Q network & \multicolumn{4}{c}{MLP(128,64)/MLP(64,64,5)}\\
Activation & \multicolumn{4}{c}{ReLU} \\
dimension of msg & $3$ & ---  & $3$ & $3$ \\
radius& $6/0.6$ & --- & $6/0.6$ & $6/0.6$ \\
\hline
\end{tabular}
\label{tb:hyperparas}
\end{table}

\subsection{CBRP Function and HCOMM Function}
We provide the CBRP and HCOMM function used in the LSC in Algorithm \ref{alg:CBRP} and \ref{alg:H-COMM}.

\begin{algorithm}[tb!]
\caption{CBRP: Cluster Based Routing Protocol}\label{alg:CBRP}
\begin{algorithmic}[1]
% \Function{CBRP}
\State {\bfseries Input:$({\mathcal{V}}_l^t,{\mathcal{V}}_h^t)$,$\{w_{1}^t, \cdots, w_{n}^t\}$,\;$\{\text{POSs}_1^{t}, \cdots,$ 
\\ \hspace{0.2in} $\text{POSs}_n^{t}\}$,\;$d$}
\State Define neighbours are distance $<d$, $T_e$ is a constant to control the max-waiting time, $V_u=\varnothing$ is the undecided nodes set and $\cal{E}=\varnothing$;
\State Each node $i$ broadcast its weight $w_i^t$ to neighbours;
\State $\sharp$ Maintain the structure
\For{$i$ is in high-level nodes set ${\mathcal{V}}_h^t$} 
    \If{there is a high-level nodes in neighbours and its weight is bigger than agent $i$}
    \State Pop node $i$ from ${\mathcal{V}}_h^t$ and append it to ${\mathcal{V}}_l^t$;
    \EndIf
\EndFor
\For{$i$ is in low-level nodes set ${\mathcal{V}}_l^t$ and no high-level node is in its neighbour}
\State Pop node $i$ from ${\mathcal{V}}_l^t$ and append it to ${\mathcal{V}}_u$;
\EndFor
\State $\sharp$ Elect high-level nodes
\For{$i$ is in ${\mathcal{V}}_u$ concurrently} 
    \If{does not receive larger weight for $T_e$}
    \State append $i$ to ${\mathcal{V}}_h^t$ and broadcast to neighbours;
    \Else
    \State Wait for the signal from high-level node for $2T_e$;
    \EndIf
    \If{received a signal from high-level node}
    \State append $i$ to ${\mathcal{V}}_l^t$;
    \Else 
    \State append $i$ to ${\mathcal{V}}_h^t$;
    % \END
    \EndIf
\EndFor
\State $\sharp$ Generate communication link
\For{$i$ in ${\mathcal{V}}_h^t$} 
    \For{$j$ in ${\mathcal{V}}_i^t$ and $j$ is neighbouring $i$}
    \State append $e_{ij}=0$ and $e_{ji}=0$ to $\cal{E}$;
    \EndFor
    \For{$j$ in ${\mathcal{V}}_h^t$}
    \State append $e_{ij}=0$ to $\cal{E}$;
    \EndFor
\EndFor 
\State Return $({\mathcal{V}}_l^t,{\mathcal{V}}_h^t,\cal{E})$
% \EndFunction
\end{algorithmic}
\end{algorithm}

\begin{algorithm}[tb!]
\caption{{\bf{HCOMM}}: Hierarchical Communication based Policy Module}\label{alg:H-COMM}
\begin{algorithmic}[1]
% \Function{HCOMM}
\State{\bfseries Input: ${\mathcal{V}}_l, {\mathcal{V}}_h, {\mathcal{E}}$;}
\State{$\sharp$ Intra-group aggregation}
\For{$v^i$ in ${\mathcal{V}}_l$}
    \For{$v^j$ in ${\mathcal{V}}_h$ and $(i\rightarrow j)$ in $\mathcal{E}$}
	\State {$e_{ij}=\phi^{enc}(v^i);$ 
	\Comment{Generate normal to central messages}}
    \EndFor
\EndFor

\For{$v_j$ in ${\mathcal{V}}_h$}
	\State{$\bar{e}_{j} = \rho (\left\{ e_{ij} \right\}_{(i\rightarrow j)\in {\cal{E}}});$\Comment{Central agents aggregate received messages}}
	\State{$v_{j}^h = \phi ( \bar{e}_{j}, v_j^l);$ \Comment{Generate cluster perception}}
\EndFor
\State $\sharp$ Inter-group sharing
\For{$v_j$ in ${\mathcal{V}}_h$}
    \For{$v_i$ in ${\mathcal{V}}_h$ and $(i\rightarrow j)$ in $\mathcal{E}$} 
	\State $e_{ij} = \phi (v_{i}^h,v_{i}^l);$\Comment{Generate central to central messages}
	\EndFor
\EndFor
\For{$v_j$ in ${\mathcal{V}}_h$}
	\State $\bar{e}_{j} = \rho (\left\{ e_{ij} \right\}_{(i\rightarrow j)\in {\cal{E}}});$ \Comment{Aggregate received central to central messages}
	\State $v_{j}^g = \phi ( \bar{e}_{j}, v_j^l);$\Comment{Obtain global perception}
\EndFor
\State $\sharp$ Intro-group sharing 
\For{$v_i$ in ${\mathcal{V}}_h$}
    \For{$v_j$ in ${\mathcal{V}}_l$ and $(i\rightarrow j)$ in $\mathcal{E}$}
	\State $e_{ij} = \phi (v_{i}^g,v_{i}^h,v_{i}^l, e_{ji})$, $\bar{e}_{j} = \rho (\left\{ e_{ij} \right\}_{(i\rightarrow j)\in {\cal{E}}});$ \Comment{Generate central to normal messages}
	\EndFor
\EndFor
\For{$v_j$ in ${\mathcal{V}}_l \cup {\mathcal{V}}_h$}
    \For{$v_i$ in ${\mathcal{V}}_h$ and  $(i\rightarrow j)$ in $\mathcal{E}$}
	\State $\bar{e}_{j} = \rho (\left\{ e_{ij} \right\}_{(i\rightarrow j)\in {\cal{E}}});$ \Comment{Aggregate received central to normal messages}
	\EndFor
	\State $v_{j}^n = \phi ( \bar{e}_{j}, v_j^l);$\Comment{Update states}
    \State $q_j=Q(v_j^l);$ 
\EndFor
\State return $q$  
% \EndFunction
\end{algorithmic}
\end{algorithm}

\end{document}